\documentclass[11pt]{article}

\usepackage[preprint]{acl}

\usepackage{times}
\usepackage{latexsym}

\usepackage[T1]{fontenc}

\usepackage[utf8]{inputenc}

\usepackage{microtype}

\usepackage{inconsolata}

\usepackage{graphicx}

\usepackage{booktabs,multirow}
\usepackage{array,tabularx}
\usepackage{placeins}
\usepackage{amsmath}
\usepackage{algorithm,algpseudocode}
\usepackage[table]{xcolor}

\title{Meet the Future Halfway: Foresight-Guided Speculative Decoding for Looped Transformers}
\hypersetup{
  pdftitle={Meet the Future Halfway: Foresight-Guided Speculative Decoding for Looped Transformers},
  pdfauthor={Ziyang Ma, Zihong Zhang, Tinghao Dong, Zuchao Li, Lefei Zhang}
}

\author{
  Ziyang Ma\textsuperscript{1}\quad
  Zihong Zhang\textsuperscript{2}\quad
  Tinghao Dong\textsuperscript{1}\quad
  Zuchao Li\textsuperscript{2}\quad
  Lefei Zhang\textsuperscript{1}\\[3pt]
  {\normalfont\small\textsuperscript{1}School of Computer Science, Wuhan University, Wuhan, China}\\
  {\normalfont\small\textsuperscript{2}School of Artificial Intelligence, Wuhan University, Wuhan, China}\\[3pt]
  {\normalfont\small\texttt{\{maziyang, zhangzihong, dongtinghao, zcli-charlie, zhanglefei\}@whu.edu.cn}}\\[5pt]
  {\normalfont\small Preprint:
  \href{https://openreview.net/forum?id=kwTFlHbCax}{12 October 2026 (OpenReview)}\quad\textbar\quad
  Revised: 30 November 2026}
}

\begin{document}
\maketitle
\begin{abstract}
Looped transformers gain effective depth through repeated use of shared layers, but sequential recurrence makes autoregressive generation expensive. Existing self-speculative pipelines overlap shallow prediction with full-depth verification, while their lookahead remains token-wise. We propose \textbf{LoopAhead}, a training-free decoding framework that uses early bonus logits to sustain longer drafts and advance them through the pipeline as blocks. An online draft automaton incorporates first-stage top-$k$ predictions before final verification completes, supplying fresh contextual candidates for compact draft trees. Shallow verification selects token sequences whose computed states advance together through the remaining loops in a block wavefront pipeline, overlapping drafting with verification. Across six models and five benchmarks, our method achieves average speedups of $3.75$--$7.97\times$ over SGLang autoregressive decoding under greedy generation. Code: \url{https://github.com/whucs21Mzy/loopahead}.
\end{abstract}

\section{Introduction}
\label{sec:introduction}

Looped transformers repeatedly apply a shared stack of blocks, increasing effective depth without proportionally increasing the parameter count. This enables extra test-time computation in latent space without requiring longer reasoning traces. Huginn~\citep{geiping2025latent} improves mathematical reasoning and code generation with additional recurrence, while Ouro~\citep{zhu2025ouro} demonstrates gains in multi-hop knowledge manipulation in controlled experiments. Parameter sharing, however, does not ensure fast decoding: generating each token still requires sequential recurrent passes and repeated access to the shared weights.

Recurrence also offers two opportunities for self-speculation. Intermediate states provide early next-token predictions, and shared weights allow different positions and recurrent depths to advance in the same core batch. LoopSpec~\citep{cho2026loopspec} and WaveFront Decoding~\citep{ha2026wavefront} exploit these properties by overlapping shallow drafting with full-depth verification across positions and recurrent stages.

A pipeline built from single-token proposals limits lookahead. Longer drafts can extend it, but repeated verification or discarded successor progress can erase their advantage. Early bonus logits, obtained from an intermediate prediction after the first group of recurrent iterations (denoted as $q_1$), provide fresh continuations before full-depth verification finishes. For looped models, their shallow-verified states must also retain progress through the remaining recurrent stages.

\begin{figure}[t]
    \centering
    \includegraphics[width=\columnwidth]{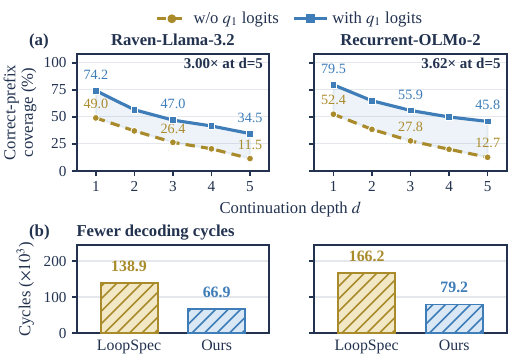}
    \caption{Early logits extend lookahead. (a) Final $d$-token prefix coverage at matched contexts and tree budget, excluding the root. (b) Full-system cycles versus LoopSpec at equal output lengths. Curves average HumanEval+, MATH500, and BBH with cycles summed.}
    \label{fig:foresight-motivation}
\end{figure}

Figure~\ref{fig:foresight-motivation}(a) shows that $q_1$-logit updates cover longer final continuations within a fixed tree budget. Realizing this lookahead requires the selected inputs to advance together rather than enter the pipeline one at a time. The full-system comparison in Figure~\ref{fig:foresight-motivation}(b) shows fewer decoding cycles than LoopSpec at matched output lengths.

We introduce \textbf{LoopAhead}, a training-free framework that turns early predictions into sustained multi-token lookahead. A request-local automaton combines historical continuations with branch-local $q_1$ top-$k$ predictions, refreshing draft candidates before full-depth verification completes. Shallow tree verification selects a contiguous sequence and reuses its computed hidden states and key-value caches as the selected inputs advance together through the remaining recurrent stages. Tail bonus predictions guide subsequent drafts while earlier blocks are still being verified. Final decisions commit outputs in order and preserve compatible successor progress, enabling multi-token commitment without an auxiliary draft model or reduced target recurrence. Implemented in SGLang, LoopAhead achieves model-wise mean greedy speedups of $3.75$--$7.97\times$ over SGLang autoregressive decoding across six models and five benchmarks, with gains also observed under stochastic sampling. Paired ablations isolate the contributions of candidate memory and recurrent execution while keeping output tokens fixed.

\section{Related Work}
\label{sec:related-work}

\subsection{Speculative Decoding and Draft Reuse}
Speculative decoding verifies inexpensive proposals in parallel and preserves the target distribution through rejection sampling, as developed by \citet{leviathan2023speculative} and \citet{chen2023speculative}. Learned approaches generate candidates through multiple decoding heads in Medusa~\citep{cai2024medusa} or feature extrapolation in EAGLE~\citep{li2024eagle}, whereas REST~\citep{he2024rest} retrieves continuations without training a drafter. Candidate reuse further exploits information already produced during decoding: SSSD~\citep{marzollo2026sssd} combines CPU-based $n$-gram retrieval from prompts, generated outputs, and an optional datastore with hardware-aware tree speculation. Token Recycling~\citep{luo2025recycling} refreshes a token-adjacency structure with model predictions, and RACER~\citep{zhang2026racer} combines logit-based candidates with contextual pattern retrieval. ECHO~\citep{ma2026echo} explores draft trees using early-layer bonus logits and cached intermediate states.

\subsection{Pipelined Self-Speculation}
Self-speculative decoding (SSD) obtains proposals within the target model: Draft \& Verify~\citep{zhang2024draftverify} skips layers during drafting, while LayerSkip~\citep{elhoushi2024layerskip} trains early exits and shares computation with later-layer verification. SPEED~\citep{hooper2023speed} uses early predictions to pipeline future tokens in weight-shared decoders. PipeSpec~\citep{mcdanel2025pipespec} instead pipelines multiple models, while Speculative Pipeline Decoding~\citep{yu2026spd} overlaps a learned draft module with target stages. For native looped models, LoopSpec~\citep{cho2026loopspec} overlaps shallow proposals with full-depth verification and selectively adds a deeper proposal. WaveFront Decoding~\citep{ha2026wavefront} batches token states across positions and recurrent depths. DAS-Wave~\citep{li2026daswave} combines mature-prefix reads with parallel refinement and an independent full-depth verifier. LoopAhead coordinates candidate updates, shallow verification, and the progression of selected token spans through the full recurrent computation.

\section{Preliminaries}
\label{sec:preliminaries}

\subsection{Standard Speculative Decoding}
\label{subsec:standard-speculation}

Speculative decoding pairs inexpensive drafting with parallel target verification (\citealp{leviathan2023speculative} and \citealp{chen2023speculative}). A drafter proposes $L$ tokens. The target scores them in parallel, but acceptance proceeds in prefix order. Let $p$ and $q$ denote normalized target and proposal distributions under the same causal context $c$. For a draw $y\sim q(\cdot\mid c)$, acceptance and rejection correction use
\begin{equation}
 \begin{aligned}
 \alpha(y\mid c) &= \min\!\left\{1,\frac{p(y\mid c)}{q(y\mid c)}\right\},\\
 p_{\mathrm{rej}}(x\mid c) &=
 \frac{[p(x\mid c)-q(x\mid c)]_+}
      {\sum_v[p(v\mid c)-q(v\mid c)]_+}.
 \end{aligned}
 \label{eq:speculative-verification}
\end{equation}
Here $[u]_+=\max(0,u)$, and $v$ ranges over the vocabulary. On rejection, sample from $p_{\mathrm{rej}}$ and discard the remaining draft suffix. Its denominator is positive whenever rejection can occur. If all $L$ proposals pass, sample a target bonus token at the extended prefix. These rules preserve the target distribution in exact arithmetic. Greedy verification instead keeps the longest prefix matching target argmax predictions, then emits the target correction or bonus.

\subsection{Looped Self-Speculation}
\label{subsec:looped-speculation}

A looped transformer applies a shared recurrent core $S$ times per token. Following LoopSpec~\citep{cho2026loopspec}, we group these iterations into $P=S/K$ stages, assuming $K\mid S$. Let $z_j(c)$ be the next-token logits after $j\times K$ iterations under context $c$, for $j=1,\ldots,P$. Define
\begin{equation}
 \begin{aligned}
 p_j(\cdot\mid c) &= \operatorname{softmax}\!\bigl(z_j(c)\bigr),\\
 g_j(c) &= \operatorname*{arg\,max}_{x} p_j(x\mid c).
 \end{aligned}
 \label{eq:recurrent-stages}
\end{equation}
The argmax $g_j$ uses a fixed tie-breaking rule. We consider two proposal endpoints: the first ($q_1$) reads $p_1$ after $K$ iterations, while the optional second ($q_2$) reads $p_X$ after $X\times K$ iterations, where $1<X<P$. Here, $q_1$ and $q_2$ index proposal endpoints rather than consecutive recurrent stages, and the final target is $p=p_P$. For sampling, readouts are temperature-scaled and filtered before verification, and $q_2$ uses a residual proposal rather than $p_X$ directly. LoopSpec pipelines token-wise proposals across stages, overlapping shallow prediction at later positions with refinement of earlier ones. Live inputs retain hidden states and stage-specific key/value (KV) caches. The operator $\Vert$ concatenates sequences or appends a token.

\subsection{Tree Verification and Bonus Logits}
\label{subsec:tree-verification}

A draft tree $\mathcal{T}$ represents alternative continuations with shared prefixes. Tree attention scores its nodes in parallel under their respective causal contexts, as in Medusa~\citep{cai2024medusa} and EAGLE~\citep{li2024eagle}. Let $c^{-}$ be the context before the root. Write $\mathcal{P}$ for its prefix-key columns and $\operatorname{Anc}^{+}(v)$ for the key columns of node $v$ and its ancestors. The additive attention mask is
\begin{equation}
 M_{\mathcal{T}}(v,u)=
 \begin{cases}
 0,&u\in\mathcal{P}\cup\operatorname{Anc}^{+}(v),\\
 -\infty,&\text{otherwise}.
 \end{cases}
 \label{eq:tree-causal-mask}
\end{equation}
A node at depth $d(v)$ uses position $|c^{-}|+d(v)$, with root depth zero. Sibling branches remain invisible to one another.
For a root-to-node path $B=(b_1,\ldots,b_m)$, write $b_{1:i}=(b_1,\ldots,b_i)$ and $c_i=c^{-}\mathbin{\Vert}b_{1:i}$. The last input's readout at stage $j$ provides the bonus logits and, under greedy decoding, the next token beyond $B$:
\begin{equation}
 z_j^{\mathrm{bonus}}=z_j(c_m),\qquad \beta_j=g_j(c_m).
 \label{eq:bonus-readout}
\end{equation}
For $j=1$, this is an early proposal. A target bonus uses $j=P$ and is emitted only after its preceding path passes final verification. Predicting beyond the bonus requires a readout with that token included in the context.

\section{Method}
\label{sec:method}

\begin{figure*}[t]
\centering
\includegraphics[width=\textwidth]{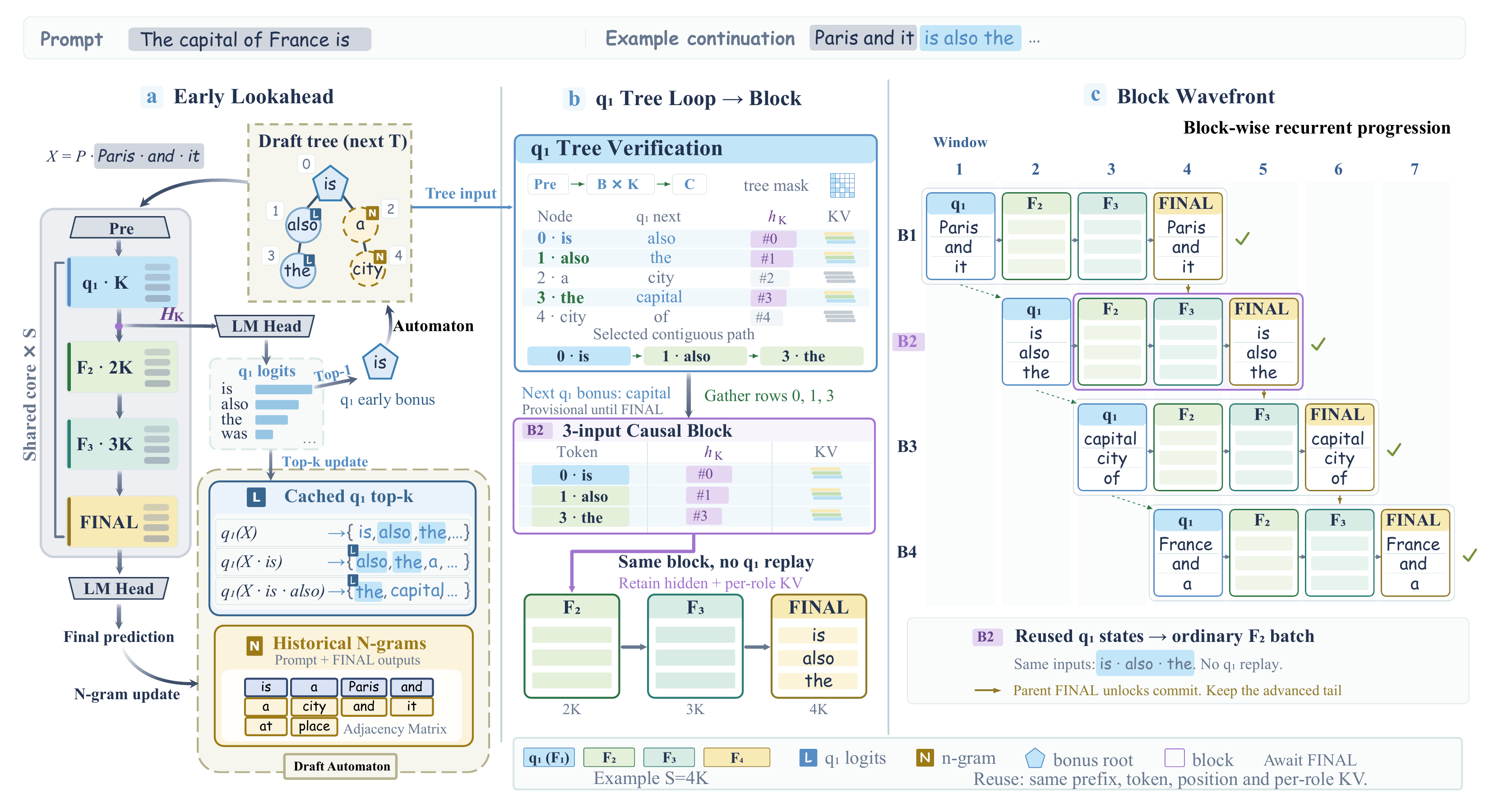}
\caption{Early-logit drafting and block-wavefront decoding ($S=4K$). (a) The preceding $q_1$ bonus roots a tree drawn from cached top-$k$ candidates (L) and historical $n$-grams (N). (b) $q_1$ selects \emph{is also the}, retains its states, and predicts bonus \emph{capital}. (c) Blocks advance toward ordered FINAL commitment while later blocks enter the pipeline. Windows show dependencies.}
\label{fig:framework}
\end{figure*}

LoopAhead uses $q_1$ to select a contiguous path from a draft tree, then advances this provisional block through the remaining loops using its computed hidden states and KV caches (Figure~\ref{fig:framework}). Early logits refresh the draft automaton while older blocks await final verification. Final predictions determine ordered commitment, and matching successors retain their progress. We preserve all $S$ loops and native input/readout modules without an auxiliary drafter. The description below follows greedy decoding.

\subsection{Early-Logit Drafting}
\label{subsec:method-updates}

To construct draft trees, we maintain a request-local Aho--Corasick automaton that combines historical $n$-gram continuations from the prompt and committed outputs with early top-$k$ predictions from $q_1$. The predictions remain provisional within their producing branch until final verification, while compatible descendants can use them to construct subsequent drafts.

For a computed readout $z_1(c)$ under context $c$, the current branch's candidate memory $\mathcal{M}_b$ is updated as
\begin{equation}
 \mathcal{M}_b[\kappa_n(c)]
 \leftarrow \operatorname{Top}_k\!\bigl(z_1(c)\bigr).
 \label{eq:q1-publication}
\end{equation}
Here $\operatorname{Top}_k$ returns an ordered list of candidate tokens, and $\kappa_n(c)$ is the suffix of at most $n$ context tokens used as the lookup key. These updates are available to descendants of the same branch as soon as the $q_1$ result is consumed, but remain provisional until final verification. When the corresponding final prediction is committed, we retain the original $q_1$ candidates in committed memory and insert the newly observed $n$-grams into history. Rejected branches discard their local updates.

The next draft tree $\mathcal{T}$ is rooted at an available input token $r$, typically the $q_1$ bonus prediction from the preceding block. We merge shared prefixes and retain a leading path with alternative branches under node budget $N$ and path-length cap $D$. Each path contains at most $D$ inputs including its root, and $|\mathcal{T}|\le N$. Trees launch when candidate evidence, prefix dependencies, and capacity permit. To avoid unproductive tree verification, we skip tree construction for roots with consistently short $q_1$-selected paths, while periodically retrying them as the candidate memory evolves. When tree verification is skipped, decoding continues through the ordinary token-wise pipeline.

\subsection{Shallow Tree Verification}
\label{subsec:method-tree}

The $q_1$ stage evaluates all draft-tree nodes through the first $K$ loops using the causal mask in Eq.~\ref{eq:tree-causal-mask}. Starting from the root $b_1=r$, we follow the child matching the $q_1$ greedy prediction at each position, forming a contiguous input block $B=(b_1,\ldots,b_m)$. Selection stops when no matching child is available or a length or capacity limit is reached. The selected path and its $q_1$ bonus $\beta=\beta_1$ satisfy
\begin{equation}
 \begin{aligned}
 b_{i+1} &= g_1(c_i), &&1\le i<m,\\
 \beta &= g_1(c_m). &&
 \end{aligned}
 \label{eq:q1-path}
\end{equation}
All $m$ selected inputs retain their $q_1$ hidden states and KV caches for subsequent recurrent stages. Their logits, including the bonus row $z_1(c_m)$, update the draft memory through Eq.~\ref{eq:q1-publication}. The bonus $\beta$ can seed further lookahead before $B$ completes final verification, but predicting beyond $\beta$ requires a readout under $c_m\mathbin{\Vert}\beta$, obtained from a compatible computed successor or a new execution. In Figure~\ref{fig:framework}(b), the block contains inputs \emph{is}, \emph{also}, and \emph{the}, and the last input predicts bonus \emph{capital}. These are provisional input states whose predictions still require FINAL verification.

\begin{figure*}[t]
\centering
\includegraphics[width=\textwidth]{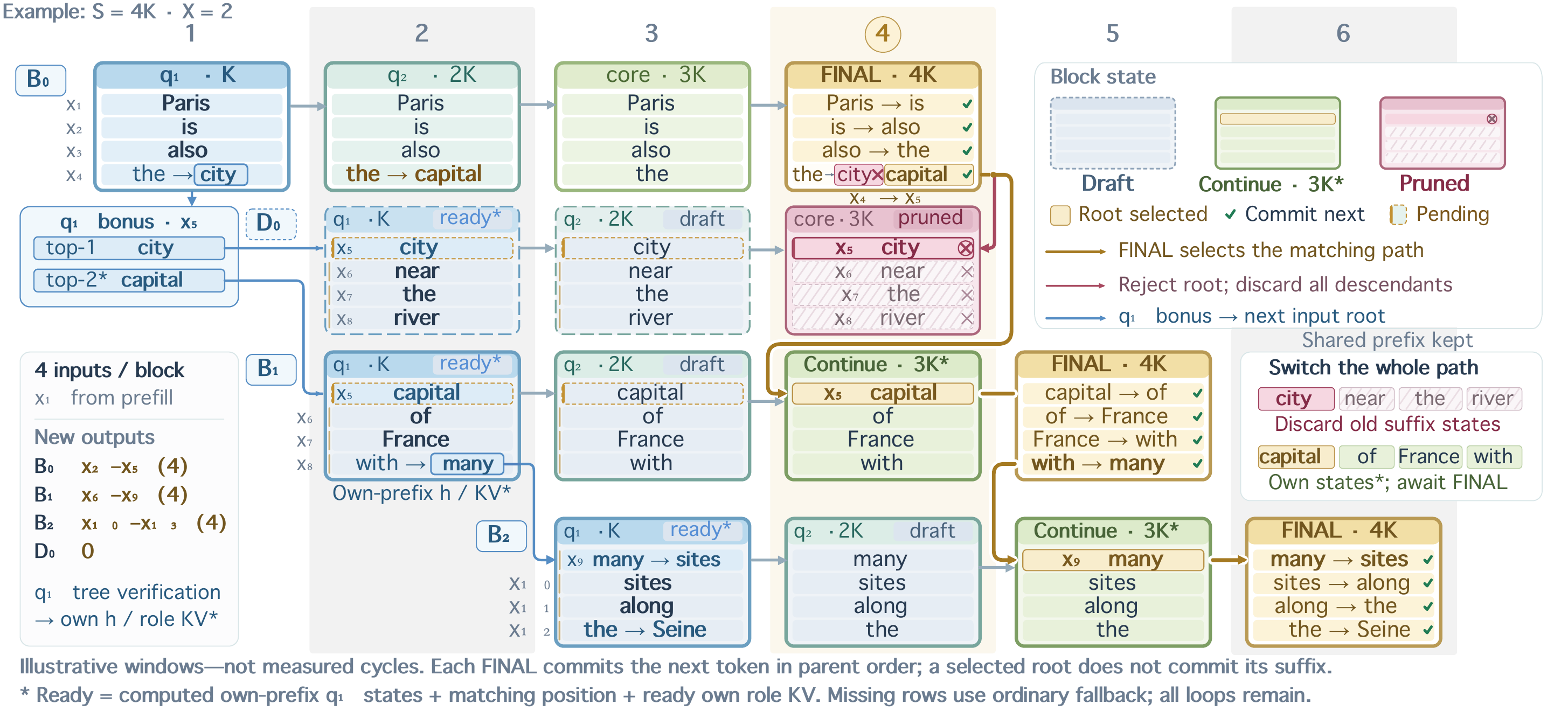}
\caption{Continuation after a rejected branch ($S=4K$, $X=2$). $B_0$'s FINAL selects \emph{capital} over \emph{city}, discarding $D_0$ and preserving compatible $B_1$ and $B_2$ progress. Starred states require matching contexts and ready hidden/KV. Suffix outputs retain their own ordered FINAL checks. Windows show dependencies.}
\label{fig:branch-pipeline}
\end{figure*}

\subsection{Block Wavefront Decoding}
\label{subsec:method-wavefront}

\paragraph{Advancing computed states.}
The selected block resumes from its cached $q_1$ states at depth $K$. At each subsequent recurrent stage, all dependency-ready inputs in the block advance together. Let $H_{j-1}$ collect their recurrent and native auxiliary states. For $j=2,\ldots,P$, stage $j$ computes
\begin{equation}
 (H_j,\mathsf{C}_j)
 =F_j\bigl(H_{j-1},\mathsf{C}^{-}_j,M_{\triangle}\bigr).
 \label{eq:block-wavefront}
\end{equation}
Here $F_j$ executes the next $K$ loops and required native modules, $\mathsf{C}^{-}_j$ contains the corresponding prefix KV, and $\mathsf{C}_j$ contains the newly written KV. These caches remain specific to each recurrent iteration and attention role. The chain mask $M_{\triangle}$ is Eq.~\ref{eq:tree-causal-mask} restricted to the selected path: input $i$ sees preceding segment inputs, itself, and the matching prefix. Prefix dependencies must be ready or supplied by compatible ancestors in the same causal batch.

Within a block, consecutive inputs advance together through each ready stage. Across blocks, the scheduler advances older inputs at later stages while new trees enter $q_1$ (Figure~\ref{fig:framework}(c)). A block's tail can seed the next tree before that block finishes verification. Supported tree and ordinary rows share core execution with their own attention contexts, and CUDA graphs remain enabled.

\paragraph{Committing a continuous prefix.}
Although the selected block advances together, its outputs must be committed in causal order. Once the block root $r$ has been committed, we emit the longest consecutive prefix supported by available FINAL predictions, stopping at the first mismatch. Let $c_1=c^-\mathbin{\Vert}(r)$ denote the committed prefix ending at the root. Let $a\le m$ count the consecutive FINAL-ready inputs from the start of $B$, with $y_i=g_P(c_i)$ for $1\le i\le a$. If $a=0$, we wait. Otherwise, the number of emitted tokens $\ell$ and the output sequence are given by
\begin{align}
 &\ell = \min\!\bigl(\{\,i:1\le i<a,\ y_i\ne b_{i+1}\,\}\cup\{a\}\bigr),\notag\\
 &\Delta_{\mathrm{out}} = (b_2,\ldots,b_{\ell},y_{\ell}).
 \label{eq:final-prefix-commit}
\end{align}
This emits $\ell$ tokens: $\ell-1$ matched successors followed by the final correction or next-token prediction at the ready boundary. When $\ell=1$, the matched part is empty. The root is not emitted again, and EOS and length limits take precedence.

For the block \emph{is also the}, suppose all three FINAL predictions are \emph{also}, \emph{the}, and \emph{capital}. Once \emph{is} belongs to the committed prefix, these three predictions can be emitted in the same cycle. If the matching \emph{capital} input has already reached depth $2K$, it continues from that depth.

\paragraph{Continuing the surviving wavefront.}
The last emitted token $y_{\ell}$ selects the successor under context $c_{\ell}\mathbin{\Vert} y_{\ell}$. We retain its computed states and compatible descendants, pruning the rejected continuation rather than restarting all successors. Reuse requires the complete context, position, and state/KV provenance to match. If no eligible successor exists, that input starts at depth zero. A successor already at FINAL permits further commitment in the same cycle. In Figure~\ref{fig:branch-pipeline}, FINAL selects \emph{capital} instead of \emph{city}. The \emph{city} continuation is discarded, while the matching \emph{capital} branch and its compatible descendants retain their progress. Each descendant still requires its own ordered FINAL decision. Algorithm~\ref{alg:block-wavefront} gives the complete schedule, with cache and sampling semantics in Appendix~\ref{app:verification-semantics}.

\paragraph{Preparing alternatives.}
Tree side branches supply candidate $q_1$ states, while pipeline alternatives advance through later stages. Following LoopSpec~\citep{cho2026loopspec}, optional $q_2$ proposes $g_X(c)$ when $g_X(c)\ne g_1(c)$. Its successor uses context $c\mathbin{\Vert} g_X(c)$, starts from a matching computed state or depth zero, and coexists with the $q_1$ successor until parent FINAL. Admitted alternatives join ordinary dependency-ready batches before selection. Prewarming is bounded and model-dependent.

\begin{table*}[!t]
\centering
\begingroup
\definecolor{LoopAheadBlue}{rgb}{0.88,0.94,1.0}
\fontsize{8.5}{10}\selectfont
\setlength{\tabcolsep}{1.2pt}
\renewcommand{\arraystretch}{1.03}
\begin{tabular*}{\textwidth}{@{}ll@{\extracolsep{\fill}}*{6}{rr}@{}}
\toprule
\multirow{2}{*}{\textbf{Model}} & \multirow{2}{*}{\textbf{Method}}
& \multicolumn{2}{c}{\textbf{HumanEval+}}
& \multicolumn{2}{c}{\textbf{MBPP+}} & \multicolumn{2}{c}{\textbf{MATH500}}
& \multicolumn{2}{c}{\textbf{BBH}} & \multicolumn{2}{c}{\textbf{GSM8K}}
& \multicolumn{2}{c}{\textbf{Average}} \\
\cmidrule(lr){3-4}\cmidrule(lr){5-6}\cmidrule(lr){7-8}
\cmidrule(lr){9-10}\cmidrule(lr){11-12}\cmidrule(lr){13-14}
& & Speedup & $\gamma$ & Speedup & $\gamma$ & Speedup & $\gamma$
& Speedup & $\gamma$ & Speedup & $\gamma$ & Speedup & $\gamma$ \\
\midrule
 & SSSD & $1.74\!\times$ & 3.43 & $1.04\!\times$ & 1.81 & $0.81\!\times$ & 2.35 & $1.42\!\times$ & 4.18 & $0.84\!\times$ & 1.97 & $1.17\!\times$ & 2.75 \\
 & NGRAM & $1.72\!\times$ & 2.21 & $2.10\!\times$ & 2.62 & $1.71\!\times$ & 2.69 & $2.11\!\times$ & 3.31 & $1.69\!\times$ & 2.39 & $1.87\!\times$ & 2.64 \\
 & RACER & $3.60\!\times$ & 5.21 & $2.12\!\times$ & 2.98 & $1.68\!\times$ & 3.25 & $2.48\!\times$ & 4.69 & $1.76\!\times$ & 2.94 & $2.33\!\times$ & 3.81 \\
 & WFD & $3.21\!\times$ & 3.69 & $2.89\!\times$ & 3.31 & $2.46\!\times$ & 3.12 & $2.60\!\times$ & 3.30 & $2.38\!\times$ & 2.86 & $2.71\!\times$ & 3.26 \\
 & LoopSpec & $3.55\!\times$ & 3.81 & $3.47\!\times$ & 3.62 & $2.45\!\times$ & 3.47 & $2.53\!\times$ & 3.58 & $2.71\!\times$ & 3.28 & $2.94\!\times$ & 3.55 \\
\rowcolor{LoopAheadBlue}[3pt][3pt]
\cellcolor{white}\multirow{-6}{*}{Ouro-1.4B} & \textbf{LoopAhead} & $\mathbf{4.94}\!\times$ & \textbf{6.97} & $\mathbf{4.08}\!\times$ & \textbf{5.18} & $\mathbf{3.10}\!\times$ & \textbf{5.22} & $\mathbf{3.67}\!\times$ & \textbf{6.58} & $\mathbf{2.95}\!\times$ & \textbf{4.26} & $\mathbf{3.75}\!\times$ & \textbf{5.64} \\
\midrule
 & SSSD & $1.74\!\times$ & 3.33 & $0.97\!\times$ & 1.55 & $0.67\!\times$ & 1.94 & $1.45\!\times$ & 3.94 & $1.44\!\times$ & 2.82 & $1.26\!\times$ & 2.72 \\
 & NGRAM & $1.83\!\times$ & 2.22 & $2.18\!\times$ & 2.50 & $1.54\!\times$ & 2.36 & $2.04\!\times$ & 2.98 & $2.09\!\times$ & 2.58 & $1.93\!\times$ & 2.53 \\
 & RACER & $3.78\!\times$ & 4.95 & $2.14\!\times$ & 2.57 & $1.67\!\times$ & 2.89 & $2.76\!\times$ & 4.47 & $2.83\!\times$ & 3.81 & $2.63\!\times$ & 3.74 \\
 & WFD & $3.45\!\times$ & 3.76 & $3.33\!\times$ & 3.44 & $2.61\!\times$ & 3.11 & $2.97\!\times$ & 3.45 & $3.14\!\times$ & 3.41 & $3.10\!\times$ & 3.43 \\
 & LoopSpec & $3.90\!\times$ & 3.85 & $3.97\!\times$ & 3.66 & $2.55\!\times$ & 3.47 & $2.89\!\times$ & 3.68 & $3.51\!\times$ & 3.64 & $3.36\!\times$ & 3.66 \\
\rowcolor{LoopAheadBlue}[3pt][3pt]
\cellcolor{white}\multirow{-6}{*}{Ouro-2.6B} & \textbf{LoopAhead} & $\mathbf{6.46}\!\times$ & \textbf{7.31} & $\mathbf{4.60}\!\times$ & \textbf{4.38} & $\mathbf{3.01}\!\times$ & \textbf{4.35} & $\mathbf{4.73}\!\times$ & \textbf{6.83} & $\mathbf{5.31}\!\times$ & \textbf{6.12} & $\mathbf{4.82}\!\times$ & \textbf{5.80} \\
\midrule
 & SSSD & $1.57\!\times$ & 3.41 & $0.96\!\times$ & 1.92 & $1.50\!\times$ & 3.93 & $1.33\!\times$ & 3.52 & $1.00\!\times$ & 2.12 & $1.27\!\times$ & 2.98 \\
 & NGRAM & $1.37\!\times$ & 2.45 & $1.41\!\times$ & 2.51 & $2.39\!\times$ & 4.39 & $1.28\!\times$ & 2.36 & $1.09\!\times$ & 1.95 & $1.51\!\times$ & 2.73 \\
 & RACER & $2.63\!\times$ & 4.71 & $1.58\!\times$ & 2.80 & $2.60\!\times$ & 4.83 & $2.15\!\times$ & 3.98 & $1.70\!\times$ & 3.04 & $2.13\!\times$ & 3.87 \\
 & WFD & $3.82\!\times$ & 5.02 & $2.77\!\times$ & 3.56 & $3.38\!\times$ & 4.74 & $2.71\!\times$ & 3.76 & $2.35\!\times$ & 3.04 & $3.01\!\times$ & 4.02 \\
 & LoopSpec & $4.92\!\times$ & 6.49 & $4.30\!\times$ & 5.51 & $4.36\!\times$ & 6.24 & $3.86\!\times$ & 5.51 & $3.82\!\times$ & 5.03 & $4.25\!\times$ & 5.76 \\
\rowcolor{LoopAheadBlue}[3pt][3pt]
\cellcolor{white}\multirow{-6}{*}{huginn-0125} & \textbf{LoopAhead} & $\mathbf{5.58}\!\times$ & \textbf{9.44} & $\mathbf{4.71}\!\times$ & \textbf{6.83} & $\mathbf{5.49}\!\times$ & \textbf{10.07} & $\mathbf{4.08}\!\times$ & \textbf{6.94} & $\mathbf{4.20}\!\times$ & \textbf{6.22} & $\mathbf{4.81}\!\times$ & \textbf{7.90} \\
\midrule
 & SSSD & $2.55\!\times$ & 3.83 & $1.68\!\times$ & 2.40 & $1.97\!\times$ & 3.23 & $2.58\!\times$ & 4.32 & $1.43\!\times$ & 2.08 & $2.04\!\times$ & 3.17 \\
 & NGRAM & $1.78\!\times$ & 3.06 & $2.14\!\times$ & 3.39 & $1.78\!\times$ & 3.69 & $1.56\!\times$ & 3.41 & $1.21\!\times$ & 2.02 & $1.69\!\times$ & 3.11 \\
 & RACER & $2.49\!\times$ & 4.72 & $2.02\!\times$ & 3.43 & $1.79\!\times$ & 4.11 & $1.95\!\times$ & 4.64 & $1.80\!\times$ & 3.17 & $2.01\!\times$ & 4.01 \\
 & WFD & $5.71\!\times$ & 11.46 & $4.49\!\times$ & 8.84 & $4.68\!\times$ & 9.74 & $4.84\!\times$ & 10.03 & $3.64\!\times$ & 7.13 & $4.67\!\times$ & 9.44 \\
 & LoopSpec & $6.64\!\times$ & 14.15 & $5.84\!\times$ & 12.44 & $6.02\!\times$ & 13.41 & $5.92\!\times$ & 13.23 & $5.17\!\times$ & \textbf{11.25} & $5.92\!\times$ & 12.89 \\
\rowcolor{LoopAheadBlue}[3pt][3pt]
\cellcolor{white}\multirow{-6}{*}{\shortstack[l]{Raven-\\Llama-3.2}} & \textbf{LoopAhead} & $\mathbf{8.42}\!\times$ & \textbf{18.47} & $\mathbf{6.98}\!\times$ & \textbf{13.36} & $\mathbf{7.68}\!\times$ & \textbf{16.33} & $\mathbf{7.21}\!\times$ & \textbf{16.15} & $\mathbf{5.59}\!\times$ & 10.75 & $\mathbf{7.18}\!\times$ & \textbf{15.01} \\
\midrule
 & SSSD & $3.29\!\times$ & 4.09 & $1.88\!\times$ & 2.22 & $2.69\!\times$ & 3.62 & $3.47\!\times$ & 4.78 & $2.27\!\times$ & 2.79 & $2.72\!\times$ & 3.50 \\
 & NGRAM & $1.82\!\times$ & 2.97 & $1.98\!\times$ & 2.83 & $1.83\!\times$ & 3.88 & $1.68\!\times$ & 3.85 & $1.42\!\times$ & 2.29 & $1.75\!\times$ & 3.16 \\
 & RACER & $3.02\!\times$ & 5.40 & $2.10\!\times$ & 3.21 & $1.88\!\times$ & 4.33 & $2.06\!\times$ & 5.04 & $2.09\!\times$ & 3.54 & $2.23\!\times$ & 4.30 \\
 & WFD & $6.80\!\times$ & 11.23 & $5.15\!\times$ & 8.35 & $5.88\!\times$ & 9.93 & $5.60\!\times$ & 9.54 & $5.12\!\times$ & 8.39 & $5.71\!\times$ & 9.49 \\
 & LoopSpec & $7.82\!\times$ & 13.96 & $6.60\!\times$ & \textbf{11.95} & $7.35\!\times$ & 13.55 & $7.08\!\times$ & 13.16 & $6.61\!\times$ & 12.13 & $7.09\!\times$ & 12.95 \\
\rowcolor{LoopAheadBlue}[3pt][3pt]
\cellcolor{white}\multirow{-6}{*}{\shortstack[l]{Recurrent-\\TinyLlama-3T}} & \textbf{LoopAhead} & $\mathbf{8.57}\!\times$ & \textbf{16.86} & $\mathbf{6.78}\!\times$ & 11.76 & $\mathbf{8.92}\!\times$ & \textbf{17.52} & $\mathbf{8.35}\!\times$ & \textbf{17.39} & $\mathbf{7.25}\!\times$ & \textbf{13.29} & $\mathbf{7.97}\!\times$ & \textbf{15.36} \\
\midrule
 & SSSD & $1.86\!\times$ & 3.99 & $1.82\!\times$ & 3.70 & $1.38\!\times$ & 3.61 & $1.62\!\times$ & 4.58 & $1.54\!\times$ & 3.28 & $1.64\!\times$ & 3.83 \\
 & NGRAM & $2.19\!\times$ & 3.72 & $2.84\!\times$ & 4.76 & $2.26\!\times$ & 4.25 & $2.27\!\times$ & 4.41 & $1.94\!\times$ & 3.29 & $2.30\!\times$ & 4.09 \\
 & RACER & $2.62\!\times$ & 4.75 & $2.51\!\times$ & 4.44 & $2.24\!\times$ & 4.58 & $2.47\!\times$ & 5.32 & $2.29\!\times$ & 4.23 & $2.43\!\times$ & 4.66 \\
 & WFD & $4.58\!\times$ & 7.47 & $4.64\!\times$ & 7.48 & $3.97\!\times$ & 7.01 & $3.77\!\times$ & 6.80 & $4.29\!\times$ & 7.04 & $4.25\!\times$ & 7.16 \\
 & LoopSpec & $6.05\!\times$ & 14.23 & $6.29\!\times$ & 14.40 & $4.89\!\times$ & 13.18 & $4.47\!\times$ & 12.80 & $5.50\!\times$ & 13.27 & $5.44\!\times$ & 13.58 \\
\rowcolor{LoopAheadBlue}[3pt][3pt]
\cellcolor{white}\multirow{-6}{*}{\shortstack[l]{Recurrent-\\OLMo-2}} & \textbf{LoopAhead} & $\mathbf{7.44}\!\times$ & \textbf{17.58} & $\mathbf{8.86}\!\times$ & \textbf{21.06} & $\mathbf{6.48}\!\times$ & \textbf{16.66} & $\mathbf{6.14}\!\times$ & \textbf{16.89} & $\mathbf{6.56}\!\times$ & \textbf{14.77} & $\mathbf{7.10}\!\times$ & \textbf{17.39} \\
\bottomrule
\end{tabular*}
\endgroup
\caption{Greedy decoding. LoopSpec uses its best tuned configuration. LoopSpec/LoopAhead results average three paired runs. Speedups are relative to SGLang AR. $\gamma$ is stage-normalized commitment. Average equally weights the five tasks. Bold marks the best result per column within each model.}
\label{tab:main-performance}
\end{table*}

\section{Experiments}
\label{sec:experiments}

\subsection{Experimental Setup}
\label{subsec:experimental-setup}

\paragraph{Datasets and Models.}
We evaluate eight checkpoints spanning the Ouro, Ouro-Thinking, and Raven-style architectures. Table~\ref{tab:main-performance} reports six non-Thinking checkpoints. The two Thinking checkpoints and their ablations are discussed separately in Appendix~\ref{app:thinking}. We evaluate code generation on HumanEval+ and MBPP+~\citep{liu2023evalplus}, mathematics on MATH500 (\citealp{hendrycks2021math} and \citealp{lightman2023verify}), and reasoning on BBH~\citep{suzgun2022bbh} and GSM8K~\citep{cobbe2021gsm8k}. All checkpoints retain their full target recurrence without additional training or an auxiliary draft model.

\paragraph{Baselines and Metrics.}
We compare with LoopSpec \citep{cho2026loopspec} on identical inputs, using the same GPU for greedy comparisons. We tune LoopSpec and select its best configuration. The comparison uses three paired runs. Appendix~\ref{app:controlled-comparison} compares them under matched recurrent and proposal schedules. Additional baselines include SGLang autoregressive decoding (AR) \citep{zheng2023sglang}, SSSD \citep{marzollo2026sssd}, SGLang's native $n$-gram speculative decoding (NGRAM), RACER \citep{zhang2026racer}, and WaveFront Decoding (WFD)~\mbox{\citep{ha2026wavefront}}.
We report decode throughput, speedups over SGLang AR, and stage-normalized commitment $\gamma=(S/K)N_{\mathrm{decode}}/C$, where $N_{\mathrm{decode}}$ is the number of generated decode tokens, $C$ is the ordinary pipeline cycle count, and $S/K$ is the number of recurrent stages. This metric measures the number of tokens committed per cycle, scaled by the recurrent stage count. Further metric details appear in Appendix~\ref{app:mechanism-counters}.
Results are averaged equally across the five benchmarks.

\paragraph{Hyperparameters and Settings.}
The main experiments run each request on one RTX 3090 in BF16 with CUDA graphs enabled. The matched-schedule comparison uses 8-node, depth-6 trees and a tree-yield gate of 2 (3 for OLMo). Sampling and ablation settings are given in Appendix~\ref{app:execution-protocol}. Generation/context limits are 1,024/2,048 tokens, with normal EOS, task stops, and checkpoint chat templates where available. The main comparison is greedy, with $T=1$ sampling evaluated separately.

\subsection{Main Results and Analysis}
\label{subsec:main-results}

\paragraph{Throughput and Pipeline Efficiency.}

LoopAhead outperforms tuned LoopSpec in all 30 greedy model--task pairs (Table~\ref{tab:main-performance}). Model-wise mean decode gains range from $1.12\times$ to $1.43\times$, with all 30 task-specific 95\% confidence intervals above 1. Speedups over SGLang AR range from $3.75\times$ to $7.97\times$ across model means. LoopAhead also exceeds WFD, SSSD, NGRAM, and RACER in every model--task comparison, with mean gains of $1.37$--$1.66\times$ over WFD. Relative to tuned LoopSpec, mean $\gamma$ rises from 3.66 to 5.80 for Ouro-2.6B and from 13.58 to 17.39 for OLMo, reflecting greater commitment per normalized pipeline cycle.

\paragraph{Effect of Sampling.}
At $T=1$, mean speedups over SGLang AR remain $2.63$--$4.36\times$. All six checkpoints outperform LoopSpec on every task, with model-wise mean gains of $1.06$--$1.37\times$. Both temperatures use matched readout settings (Table~\ref{tab:temperature-summary}). Complete results and sampling controls appear in Appendix~\ref{app:sampling-results}.

\begin{table}[t]
\centering
\begingroup
\small
\setlength{\tabcolsep}{2pt}
\renewcommand{\arraystretch}{1}
\begin{tabular*}{\columnwidth}{@{}l@{\extracolsep{\fill}}rrrr@{}}
\toprule
\multirow{2}{*}{Model} & \multicolumn{2}{c}{$T=0$} & \multicolumn{2}{c}{$T=1$} \\
\cmidrule(lr){2-3}\cmidrule(lr){4-5}
& LS & LoopAhead & LS & LoopAhead \\
\midrule
Ouro-1.4B & $2.50\!\times$ & $\mathbf{3.76}\!\times$ & $2.18\!\times$ & $\mathbf{2.63}\!\times$ \\
Ouro-2.6B & $2.84\!\times$ & $\mathbf{4.77}\!\times$ & $2.53\!\times$ & $\mathbf{3.46}\!\times$ \\
huginn-0125 & $3.20\!\times$ & $\mathbf{4.81}\!\times$ & $3.02\!\times$ & $\mathbf{3.46}\!\times$ \\
Raven-Llama-3.2 & $3.14\!\times$ & $\mathbf{4.89}\!\times$ & $3.05\!\times$ & $\mathbf{3.23}\!\times$ \\
TinyLlama-3T & $3.36\!\times$& $\mathbf{5.30}\!\times$ & $3.32\!\times$ & $\mathbf{3.57}\!\times$ \\
OLMo-2 & $4.46\!\times$ & $\mathbf{6.80}\!\times$ & $3.91\!\times$ & $\mathbf{4.36}\!\times$ \\
\bottomrule
\end{tabular*}
\endgroup
\caption{Temperature comparison: mean speedups over SGLang AR across five tasks. LS denotes LoopSpec. OLMo uses $q_1/q_2=4/16$ at both temperatures. Bold marks the faster method at each temperature.}
\label{tab:temperature-summary}
\end{table}

\subsection{Cross-Hardware Evaluation}
\label{subsec:hardware-comparison}

\begin{figure}[t]
    \centering
    \includegraphics[width=\columnwidth]{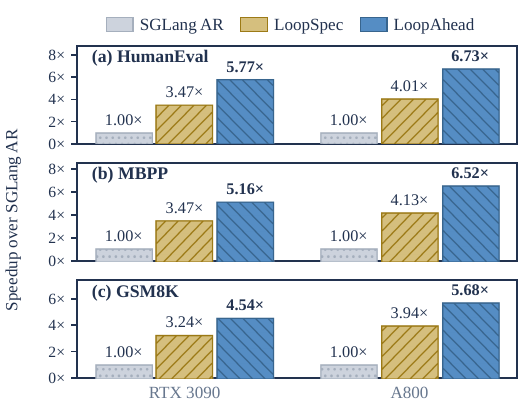}
    \caption{Mean greedy speedups over each GPU's SGLang AR with EOS truncation (six non-Thinking checkpoints per task, task-specific vertical scales).}
    \label{fig:hardware-comparison}
\end{figure}

Figure~\ref{fig:hardware-comparison} compares AR-normalized throughput across six non-Thinking checkpoints and three shared benchmark families under matched recurrent schedules. Our mean speedup is $6.31\times$ on A800 and $5.16\times$ on RTX 3090. LoopSpec reaches $4.03\times$ and $3.39\times$, respectively.  Both hardware summaries use EOS-truncated outputs. Appendix~\mbox{\ref{app:a800}} reports the A800 results.

\begin{table}[!ht]
\centering
\begingroup
\small
\setlength{\tabcolsep}{2pt}
\renewcommand{\arraystretch}{1}
\begin{tabular*}{\columnwidth}{@{}l@{\extracolsep{\fill}}rrr@{}}
\toprule
\multirow{2}{*}{Model} & \multirow{2}{*}{\shortstack{Agreement\\(\%)}} & \multicolumn{2}{c}{Non-root inputs/call} \\
\cmidrule(lr){3-4}
& & $q_1$-selected & Committed \\
\midrule
Ouro-1.4B & 91.18 & 1.36 & 1.02 \\
Ouro-2.6B & 93.13 & 1.70 & 1.35 \\
huginn-0125 & 95.77 & 1.88 & 1.68 \\
Raven-Llama-3.2 & 99.59 & 2.05 & 2.01 \\
TinyLlama-3T & 99.61 & 2.03 & 1.97 \\
OLMo-2 & 98.71 & 2.83 & 2.62 \\
\bottomrule
\end{tabular*}
\endgroup
\caption{Early predictions and final tree yield under greedy decoding, pooling counters across five tasks. Agreement compares $q_1$ and FINAL on committed contexts. Both yield columns exclude the root. OLMo uses $q_1/q_2=4/16$.}
\label{tab:q1-final-yield-main}
\end{table}

\subsection{Analysis of Early-Layer Foresight}
\label{subsec:early-to-final}

Table~\ref{tab:q1-final-yield-main} separates $q_1$/final agreement from the number of tree inputs committed in the diagnostic runs. Both yield columns exclude the root, measuring lookahead beyond the already available input. For Raven-Llama, almost all $q_1$-selected non-root inputs survive to final commitment, averaging 2.01 inputs per call out of 2.05 selected. OLMo realizes 2.62 of 2.83 selected inputs. Ouro retains a smaller fraction, showing why local acceptance alone is insufficient. These surviving inputs provide reusable computation for block advancement, but reducing decoding cycles also requires them to advance and commit together.

\begin{figure}[!ht]
    \centering
    \includegraphics[width=\columnwidth]{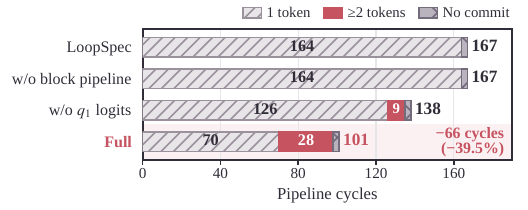}
    \caption{Cycle composition for an identical output on Raven-Llama (GSM8K).}
    \label{fig:cycle-case-study}
\end{figure}

\paragraph{Case Study: Cycle Reduction and Cost.}
For the identical output in Figure~\ref{fig:cycle-case-study}, Full commits 94 tokens in 28 multi-token cycles, reducing 167 cycles to 101 (39.5\% fewer). Without block pipelining, 66 non-root tree inputs reach FINAL, but the cycle count remains that of LoopSpec. Removing $q_1$ logit updates lowers final non-root tree reuse from 66 to 29 and raises the total to 138 cycles. Early logits improve lookahead, and block pipelining converts it into fewer pipeline cycles.

\begin{figure}[!ht]
    \centering
    \includegraphics[width=\columnwidth]{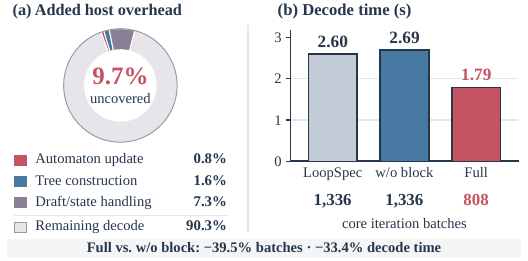}
    \caption{Uncovered host-management overhead (five repeats) and decode time for Raven-Llama on GSM8K.}
    \label{fig:case-cost-study}
\end{figure}

For the same case, automaton updates, tree construction, and draft/state handling together account for 9.7\% of decode time as host-management overhead not covered by GPU graph execution (Figure~\ref{fig:case-cost-study}). Block advancement reduces core batches from 1,336 to 808 and decode time from 2.69\,s to 1.79\,s relative to w/o block, a 33.4\% time reduction.

\subsection{Ablation Study}
\label{subsec:ablation}

\begin{table}[!ht]
\centering
\begingroup
\small
\setlength{\tabcolsep}{1pt}
\renewcommand{\arraystretch}{1}
\begin{tabular*}{\columnwidth}{@{}l@{\extracolsep{\fill}}cccc@{}}
\toprule
Model & Full & \shortstack{w/o\\block} & \shortstack{w/o\\history}
& \shortstack{w/o $q_1$\\early logits} \\
\midrule
Ouro-1.4B & $\mathbf{3.76}\!\times$ & $2.58\!\times$ & $3.42\!\times$ & $3.51\!\times$ \\
Ouro-2.6B & $\mathbf{4.77}\!\times$ & $3.16\!\times$ & $4.17\!\times$ & $4.25\!\times$ \\
huginn-0125 & $\mathbf{4.81}\!\times$ & $3.15\!\times$ & $4.59\!\times$ & $4.23\!\times$ \\
Raven-Llama-3.2 & $\mathbf{4.89}\!\times$ & $3.00\!\times$ & $4.63\!\times$ & $4.15\!\times$ \\
TinyLlama-3T & $\mathbf{5.30}\!\times$ & $3.22\!\times$ & $5.00\!\times$ & $4.54\!\times$ \\
OLMo-2 & $\mathbf{6.80}\!\times$ & $4.40\!\times$ & $6.64\!\times$ & $5.83\!\times$ \\
\bottomrule
\end{tabular*}
\endgroup
\caption{Greedy component comparisons at $T=0$. Speedups over SGLang AR, averaged equally over five tasks. Full and the separately measured controls share readout depths; OLMo uses $4/16$. The $q_1$ control disables candidate-memory updates and retains shallow verification.}
\label{tab:component-ablation}
\end{table}

Table~\ref{tab:component-ablation} compares Full with three component controls over the six main checkpoints and five tasks, with proposal depths fixed within each model.

\paragraph{Multi-Token Advancement.}
Advancing the $q_1$-selected inputs together reuses their shallow computation through later stages. The control retains early-logit updates and advances inputs individually. Full achieves $1.46$--$1.65\times$ higher mean throughput.

\paragraph{$q_1$ Early Bonus.}
Computed $q_1$ top-$k$ rows, including bonus logits, refresh draft memory for continued exploration before FINAL. Full is 7.0--18.0\% faster than removing these updates. 

\paragraph{Historical $n$-grams.}

Historical $n$-grams improve throughput by $1.69$--$2.62\%$ with both other components active (Holm-adjusted $p<0.02$, Table~\ref{tab:paired-components-3090}). Their gain reaches $1.24$--$1.27\times$ with advancement retained and the $q_1$ cache disabled.

\begin{table}[!t]
\centering
\begingroup
\small
\setlength{\tabcolsep}{1pt}
\renewcommand{\arraystretch}{1.12}
\begin{tabular*}{\columnwidth}{@{}l@{\extracolsep{\fill}}ccc@{}}
\toprule
 Model & $q_1$ cache & \shortstack{Historical\\$n$-grams}
& \shortstack{Multi-token\\advancement} \\
\midrule
 Raven-Llama & \shortstack{$1.08\!\times$\\{\scriptsize [1.07, 1.09]}} & \shortstack{$1.03\!\times$\\{\scriptsize [1.01, 1.04]}} & \shortstack{$1.39\!\times$\\{\scriptsize [1.37, 1.40]}} \\
TinyLlama & \shortstack{$1.10\!\times$\\{\scriptsize [1.09, 1.11]}} & \shortstack{$1.02\!\times$\\{\scriptsize [1.02, 1.03]}} & \shortstack{$1.40\!\times$\\{\scriptsize [1.38, 1.42]}} \\
OLMo-2 & \shortstack{$1.19\!\times$\\{\scriptsize [1.19, 1.19]}} & \shortstack{$1.02\!\times$\\{\scriptsize [1.01, 1.02]}} & \shortstack{$1.61\!\times$\\{\scriptsize [1.60, 1.62]}} \\
\bottomrule
\end{tabular*}
\endgroup
\caption{Conditional component gains, averaged geometrically over five benchmarks and three paired runs, with pointwise 95\% confidence intervals. Each component is enabled with the other two active. All nine gains have Holm-adjusted $p<0.02$.}
\label{tab:paired-components-3090}
\end{table}

\paragraph{Paired Component Controls.}
On Raven-Llama, TinyLlama, and OLMo, three paired runs of all eight settings yield identical tokens. Each component improves throughput with the other two active, and their combined gain is $1.35$--$1.51\times$ over disabling all three. Appendix~\ref{app:paired-component-controls} gives the protocol and full matrix.

\begin{table}[!ht]
\centering
\begingroup
\small
\setlength{\tabcolsep}{2pt}
\renewcommand{\arraystretch}{0.9}
\begin{tabular*}{\columnwidth}{@{}l@{\extracolsep{\fill}}cccc@{}}
\toprule
\multirow{2}{*}{Structure $(N,D)$} & \multicolumn{2}{c}{Ouro-2.6B}
& \multicolumn{2}{c}{Raven-Llama} \\
\cmidrule(lr){2-3}\cmidrule(lr){4-5}
& Speedup & $\gamma$ & Speedup & $\gamma$ \\
\midrule
Default $(8,6)$ & $4.77\!\times$ & $5.83$ & $4.89\!\times$ & $7.29$ \\
\midrule
Chain $(8,8)$ & $4.72\!\times$ & $5.79$ & $4.81\!\times$ & $6.80$ \\
Wide $(8,4)$ & $4.58\!\times$ & $5.58$ & $4.80\!\times$ & $6.94$ \\
Deep $(16,12)$ & $4.87\!\times$ & $6.25$ & $4.99\!\times$ & $7.62$ \\
Chain $(16,16)$ & $4.55\!\times$ & $5.88$ & $4.78\!\times$ & $7.06$ \\
Tree $(32,24)$ & $4.36\!\times$ & $6.14$ & $4.89\!\times$ & $7.86$ \\
\bottomrule
\end{tabular*}
\endgroup
\caption{Draft structures, averaged equally over five tasks. Speedups are relative to SGLang AR. Node budget $N$ and maximum path length $D$ include the root.}
\label{tab:tree-structure-ablation}
\end{table}

\begin{samepage}
\paragraph{Draft Structure.}
Table~\ref{tab:tree-structure-ablation} compares six draft structures on Ouro-2.6B and Raven-Llama-3.2. The 16-node, depth-12 tree has throughput close to the default 8-node, depth-6 configuration. Both single-chain variants and the wide 8-node tree have lower throughput than the displayed default. Expanding to 32 nodes increases $\gamma$ relative to this default while speedup falls to $4.36\times$ on Ouro and remains $4.89\times$ on Raven. We therefore retain the compact 8-node, depth-6 configuration as the controlled reference.

\end{samepage}

\section{Conclusion}
\label{sec:conclusion}

We presented a training-free approach for looped transformers that uses early predictions to refine future drafts and advance multiple tokens through the remaining loops together. Across the six main checkpoints and five benchmarks, model-wise mean speedups over SGLang autoregressive decoding are $3.75$--$7.97\times$ under greedy decoding and $2.63$--$4.36\times$ at $T=1$.

\section*{Limitations}

The controlled reference uses a fixed compact tree budget, although the length of reliable drafts varies with the model and context. Adapting this budget to early predictions may further improve efficiency.

\section*{Ethical Considerations}

Faster inference may make looped transformers more accessible, but it can also lower the cost of generating harmful or misleading content. Our method does not address biases or unsafe behavior in the underlying models. Deployment should retain safety checks and access controls. We evaluate throughput. Energy use and carbon emissions require separate workload-matched power measurements.

\clearpage
\appendix
\raggedbottom
\setcounter{dbltopnumber}{2}
\renewcommand{\dbltopfraction}{0.95}
\renewcommand{\textfraction}{0.05}
\makeatletter
\setlength{\@fptop}{0pt}
\setlength{\@dblfptop}{0pt}
\makeatother

\newenvironment{appendixwide}{  \begin{minipage}{\textwidth}  \renewenvironment{table*}[1][]{\captionsetup{type=table}}{}}{\end{minipage}}
\newenvironment{appendixcolumn}{  \par\addvspace{\smallskipamount}\noindent
  \begin{minipage}{\columnwidth}  \renewenvironment{table}[1][]{\captionsetup{type=table}}{}}{\end{minipage}\par\addvspace{\smallskipamount}}

\section{Datasets}
\label{app:datasets}

We evaluate code generation, mathematical problem solving, and multi-step reasoning on five benchmarks.

\paragraph{HumanEval+ and MBPP+.}
HumanEval+ extends HumanEval with additional tests for Python code generation~\citep{liu2023evalplus}. MBPP+ applies the EvalPlus framework to MBPP.\footnote{\url{https://github.com/evalplus/evalplus/releases/tag/v0.2.0}} The former uses function-completion specifications, while the latter uses short natural-language programming problems.

\paragraph{MATH500.}
MATH500 is a 500-problem subset of the MATH benchmark, covering competition-style mathematics across topics and difficulty levels (\citealp{hendrycks2021math} and \citealp{lightman2023verify}).

\paragraph{BBH.}
BIG-Bench Hard contains challenging reasoning tasks spanning logical, symbolic, and language-based problems~\citep{suzgun2022bbh}. Its varied task formats complement the more specialized code and mathematics benchmarks.

\paragraph{GSM8K.}
GSM8K contains grade-school mathematical word problems that require multi-step arithmetic reasoning~\citep{cobbe2021gsm8k}. We use its chain-of-thought evaluation variant.

\section{Additional Experimental Details}
\label{app:experimental-details}

\subsection{Checkpoints and Recurrent Stages}

{\raggedright
The shortened Recurrent-TinyLlama-3T and Recurrent-OLMo-2 labels denote \texttt{Recurrent-TinyLlama-3T-\allowbreak{}train-recurrence-32} and \texttt{Recurrent-OLMo-2-0425-\allowbreak{}train-recurrence-32}, respectively.
\par}
All benchmark comparisons retain their adapters' complete prelude, recurrent core, and output-stage computation, where applicable. The controlled comparison uses $(S,K)=(4,1)$ for Ouro and $(32,8)$ for all other models, giving four recurrent pipeline stages throughout. These counts denote recurrent stages, not physical Transformer blocks or attention layers.

\begin{samepage}
\subsection{Hardware Environment}
\label{app:hardware-environment}

Experiments use a single GPU per request without tensor parallelism. The RTX 3090 platform pairs 24\,GiB of GPU memory with dual Intel Xeon Platinum 8276L CPUs (56 physical cores total). The second platform pairs an NVIDIA A800 80GB PCIe GPU with dual Intel Xeon Platinum 8336C CPUs (64 physical cores total).

\end{samepage}

\begin{samepage}
\subsection{Software Versions}
\label{app:software-versions}

Our controlled reference implementation uses Python 3.11.5, PyTorch 2.11.0 with its CUDA 13.0 build, SGLang 0.5.14, Triton 3.6.0, and Transformers 5.8.1. Task definitions and few-shot prompts are provided by lm-evaluation-harness 0.4.12. Attention uses the Triton backend.

\end{samepage}

\subsection{Execution and Generation Protocol}
\label{app:execution-protocol}

In the matched-schedule comparison, $q_1$ and $q_2$ are read after $S/4$ and $S/2$ loops, with $K=S/4$. LoopAhead uses 8-node, depth-6 trees and a gate of 2, except for OLMo's gate of 3. Its automaton combines historical token patterns with $q_1$-logit updates. Accepted $q_1$ inputs enter the ordinary causal wavefront and finish the remaining target loops before commitment. Sampling and single-component comparisons use a common reference with OLMo at $q_1/q_2=4/16$. Appendix~\ref{app:paired-component-controls} gives the three-factor settings. Parameter sweeps specify their settings locally.

In the main experiment, requests are processed sequentially on one GPU in BF16, without tensor parallelism. Paired arms use the same GPU, tokenized inputs, full target recurrence, and stopping rules. Arm order alternates across model--task pairs. CUDA graphs remain enabled. Raven-style comparisons and within-model component ablations use shared numerical kernels.

Generation is capped at 1,024 tokens within a total context capacity of 2,048 tokens. Normal EOS and task stop strings are respected. Overlong prompts are left-truncated to 1,023 tokens to reserve the generation budget and one serving slot. We use checkpoint-provided chat templates when available, and raw completion prompts with tokenizer BOS handling for base models. Throughput measurements include all selected inputs. The LoopSpec/LoopAhead results in Table~\ref{tab:main-performance} average three paired runs. The controlled OLMo comparison in Appendix~\ref{app:controlled-comparison} uses 32 inputs per task and two paired runs. Task quality is evaluated in Appendix~\ref{app:sanity-checks}.

Tables~\ref{tab:component-ablation}, \ref{tab:thinking-component-ablation}, and~\ref{tab:q1-update-ablation-full} compare the ablation Full references with separately measured component controls. Reported cross-run gains are ratios of unrounded five-task mean throughput.

The block-pipeline ablation disables multi-input causal admission and bulk final commitment while retaining $q_1$ tree verification, logit updates, and historical retrieval. The $n$-gram ablation removes historical token-pattern seeding, insertion, and retrieval while retaining $q_1$-context indexing, suffix fallback, $q_1$-logit updates, and the block pipeline.

\subsection{Baseline Configurations}
\label{app:baseline-configurations}
The following configurations specify the baselines in Table~\ref{tab:main-performance}.
We evaluate SSSD, NGRAM, and RACER under greedy decoding with the following configurations. LoopSpec's controlled and sampling comparisons use the matched proposal schedules in Appendix~\ref{app:execution-protocol}.

SGLang AR runs all target loops for each output token ($S=4$ for Ouro and $S=32$ for the other models). 

For Table~\ref{tab:main-performance}, we sweep LoopSpec's $K$ and $q_2$ readout depth, including $q_1$-only decoding, and select one configuration per model that maximizes the equally weighted mean calibration decode speedup over AR across the five benchmarks. The selected $(K,q_1,q_2)$ depths are $(1,1,2)$ for both Ouro checkpoints, $(4,4,12)$ for Huginn, and $(2,2,8)$ for Raven-Llama, TinyLlama, and OLMo-2.

SSSD, NGRAM, and RACER each use a 64-node verification budget, including the root (at most 63 non-root candidates). SSSD retrieves from each request's prompt and committed output through the official C++ Reader, with query length 4, cached branch length 8, at most 8 children per node, and search-entry limit 100. Static and live datastores are disabled.

SGLang's native NGRAM uses breadth-first matching with breadth bounds 1 and 10, maximum history-trie depth 18, and capacity $10^7$ nodes. Each server starts with an empty trie and retains online token history across requests. External-corpus loading is disabled.

RACER combines each request's prompt and committed output with cached target logits. We set $n=10$, copy-logit top-$k=9$, history window 4,096 tokens, automaton capacity 10,000 nodes, and minimum matched suffix length 2. Logit expansion fills the remaining tree budget. Logits from all verified nodes, including nodes on rejected branches, update the candidate table; prompt-logit warmup is disabled.

WFD uses full-depth verification, with $K=1$ for Ouro, $K=2$ for Raven-Llama and TinyLlama, and $K=4$ for Huginn and OLMo-2.

\subsection{Commitment and Work Accounting}
\label{app:mechanism-counters}

For each model--task pair, throughput is $v=N_{\mathrm{decode}}/t_{\mathrm{decode}}$, using total final decode tokens and total decoding time across requests. Prefill and its first generated token are excluded. Task means give equal weight to the five benchmarks. Direct mean speedups average task-specific throughput ratios. Table~\mbox{\ref{tab:main-performance}} averages unrounded task metrics. LoopSpec/LoopAhead gains average the five paired task ratios within each run, then the three runs. Confidence intervals use Student's $t$ with two degrees of freedom.

The effective commit length is $L_{\mathrm{commit}}=N_{\mathrm{decode}}/C$. With $P=S/K$, its stage-normalized form is
\begin{equation}
  \gamma_{\mathrm{cyc}} = P\,\frac{N_{\mathrm{decode}}}{C}.
  \label{eq:gamma-cycle}
\end{equation}
All evaluated $S/K$ are integral. The main table abbreviates this as $\gamma$. $C$ counts ordinary pipeline cycles, including fill, drain, and cycles without a commitment. Additional tree and repair graph executions contribute to decode time. LoopSpec measures accepted length by committed-token source~\citep{cho2026loopspec}:
\begin{equation}
 \begin{aligned}
 \gamma_{\mathrm{LS}} &=
 \frac{S\,N_{\mathrm{decode}}}{K\,A_1+XK\,A_2+S\,A_0},\\
 A_0 &= N_{\mathrm{decode}}-A_1-A_2.
 \end{aligned}
 \label{eq:gamma-source}
\end{equation}
Here $A_1,A_2,A_0$ count mutually exclusive committed-token sources: $q_1$, $q_2$, and final fallback. Thus $\gamma_{\mathrm{LS}}\le P$, whereas $\gamma_{\mathrm{cyc}}$ can exceed $P$ through multi-token commitment. FINAL execution counters also include uncommitted branches.

For the non-pipelined speculative decoding baselines SSSD, NGRAM, and RACER~\citep{zhang2026racer}, let $V$ denote the reported target verification count and $M$ the number of requests. Their reported mean acceptance length $\bar L$ includes the first generated token of each request, which our decode-token convention excludes. The aligned serial metric is
\begin{equation}
 \gamma_{\mathrm{ser}} = \frac{N_{\mathrm{decode}}}{V}
 = \bar L-\frac{M}{V}.
 \label{eq:racer-serial-gamma}
\end{equation}
With one complete target recurrence per verification, the equivalent serial cycle count $C_{\mathrm{ser}}=PV$ gives $\gamma_{\mathrm{ser}}=PN_{\mathrm{decode}}/C_{\mathrm{ser}}$.

For WFD, the saved counter $R$ counts individual recurrent iterations, giving $\gamma_{\mathrm{WFD}}=S N_{\mathrm{decode}}/R$ under the same stage normalization.

Let $M_{\mathrm{tree}}$ denote the number of $q_1$ tree invocations, and let $N_{\mathrm{tree,final}}$ count their non-root input nodes that ultimately reach ordered final commitment. We define
\begin{equation}
  Y_{\mathrm{final}} =
  \frac{N_{\mathrm{tree,final}}}{M_{\mathrm{tree}}}.
\end{equation}
In contrast, $L_{q_1}$ counts locally accepted nodes, including the root, per invocation. The $q_1$/final agreement $\rho_1$ compares $q_1$ top-1 with the final decision only on committed contexts, using the summed matched and compared counters. Table~\ref{tab:q1-final-yield} pools raw counters across tasks before computing each ratio. Agreement $\rho_1$ and local acceptance $L_{q_1}$ characterize shallow predictions, while $Y_{\mathrm{final}}$ measures surviving non-root inputs per tree call.

\begin{table}[t]
\centering
\begingroup
\small
\setlength{\tabcolsep}{2pt}
\renewcommand{\arraystretch}{1.15}
\begin{tabular*}{\columnwidth}{@{}l@{\extracolsep{\fill}}rrr@{}}
\toprule
\textbf{Model} & $\rho_1$ (\%) & $L_{q_1}$ & $Y_{\mathrm{final}}$ \\
\midrule
Ouro-1.4B & 91.18 & 2.36 & 1.02 \\
Ouro-2.6B & 93.13 & 2.70 & 1.35 \\
Ouro-1.4B-Thinking & 86.44 & 2.52 & 0.92 \\
Ouro-2.6B-Thinking & 90.45 & 2.45 & 1.01 \\
huginn-0125 & 95.77 & 2.88 & 1.68 \\
Raven-Llama-3.2 & 99.59 & 3.05 & 2.01 \\
Recurrent-TinyLlama-3T & 99.61 & 3.03 & 1.97 \\
Recurrent-OLMo-2 & 98.71 & 3.83 & 2.62 \\
\bottomrule
\end{tabular*}
\endgroup
\caption{Greedy mechanism counters pooled over all five tasks. $\rho_1$ is $q_1$/final agreement on committed contexts. $L_{q_1}$ is local tree acceptance including the root. $Y_{\mathrm{final}}$ counts ultimately committed non-root tree inputs per call.}
\label{tab:q1-final-yield}
\end{table}

For throughput, all draft construction, graph execution, state staging, decision readback, and commitment work inside the measured decode interval is retained. We report cycles per final token, $C/N_{\mathrm{decode}}$, to account for varying output lengths under normal EOS. Output texts, token IDs, and per-request counters are retained in the source evaluation records.

\subsection{Concurrent Request Serving}
\label{app:concurrent-serving}
We evaluate four and eight concurrent clients on A800 using a mixed workload from the five benchmarks at $T=0$, with the main generation and context limits. AR uses native continuous batching. LoopAhead runs independent replicas under NVIDIA MPS, using eight for Ouro-1.4B, four for Raven-Llama and TinyLlama, and two for Huginn and OLMo-2. Each AR/LoopAhead pair uses the same physical GPU. Aggregate throughput divides total decode tokens by replay wall time, excluding startup and warmup.

\begin{table}[!t]
\centering
\begingroup
\small
\setlength{\tabcolsep}{3pt}
\renewcommand{\arraystretch}{1.12}
\begin{tabular*}{\columnwidth}{@{}l@{\extracolsep{\fill}}rrrr@{}}
\toprule
\multirow{2}{*}{Model} & \multicolumn{2}{c}{4 clients} & \multicolumn{2}{c}{8 clients} \\
\cmidrule(lr){2-3}\cmidrule(lr){4-5}
 & Speedup & P95 $\downarrow$ & Speedup & P95 $\downarrow$ \\
\midrule
Ouro-1.4B & $3.25\!\times$ & 72.2 & $2.27\!\times$ & 59.2 \\
Huginn & $1.77\!\times$ & 72.1 & $1.15\!\times$ & 56.3 \\
Raven-Llama & $4.11\!\times$ & 72.6 & $2.30\!\times$ & 57.7 \\
TinyLlama & $4.52\!\times$ & 76.2 & $2.49\!\times$ & 61.4 \\
OLMo-2 & $3.24\!\times$ & 69.9 & $1.82\!\times$ & 51.6 \\
\bottomrule
\end{tabular*}
\endgroup
\caption{Concurrent serving. Speedup compares mean aggregate decode throughput with batched AR at the same client count. P95 $\downarrow$ is the relative reduction in mean P95 end-to-end latency (\%).}
\label{tab:concurrent-serving}
\end{table}

At eight clients, all five evaluated models achieve $1.15$--$2.49\times$ the throughput of batched AR, while P95 end-to-end latency falls by $51.6$--$61.4\%$ (Table~\ref{tab:concurrent-serving}). Four-client gains reach $4.11\times$ for Raven-Llama and $4.52\times$ for TinyLlama. LoopAhead improves both aggregate throughput and tail latency at four and eight concurrent clients.

For Ouro-1.4B under Poisson arrivals at 0.1, 0.3, and 0.6 requests/s, P95 end-to-end latency falls by $80.0$--$81.1\%$.

\subsection{Draft-Structure Controls}
\label{app:tree-structure}

Table~\ref{tab:tree-structure-ablation} reports six draft configurations on two checkpoints and five tasks. The default uses $(N,D)=(8,6)$, reserve 2, and a tree-yield gate of 2, with $q_1$ expansion breadth 2 for Ouro-2.6B and 4 for Raven-Llama-3.2. The 8-node chain uses $(N,D)=(8,8)$, breadth 1, and reserve 0. The 16-node chain uses $(16,16)$ without side branches. The wide, shallow tree uses $(8,4)$, breadth 4, and reserve 4. The larger, deeper tree uses $(16,12)$ and reserve 4, retaining the default model-specific breadths. The scaled tree uses $(32,24)$ and reserve 8 with the same breadths. Both $N$ and $D$ count the root. Breadth limits $q_1$-logit expansion, while reserve controls allocation outside the preferred chain rather than imposing a hard cap on all side nodes.

Ouro-2.6B retains $(S,K,q_2)=(4,1,2)$, and Raven-Llama-3.2 retains $(32,8,16)$. Means give equal weight to tasks, and speedups use the same SGLang AR baselines as the main table. The comparison varies node budget and shape jointly. 

\subsection{Complete \texorpdfstring{$q_1$}{q1} Early-Bonus Ablation}
\label{app:q1-update-ablation}

Table~\ref{tab:q1-update-ablation-full} gives the complete per-task results for the $q_1$ early-bonus ablation, with OLMo's readout depths fixed at $4/16$.

The w/o $q_1$ early logits control removes logit-derived candidates from draft memory while retaining committed token patterns, tree verification, block scheduling, and the full target recurrence. Full incorporates computed $q_1$ top-$k$ predictions, including bonus logits.

OLMo mechanism diagnostics show that $q_1$ early-bonus updates increase continuous tree coverage from 50.1\% to 75.2\% of decode positions and final non-root yield from 2.03 to 2.63 inputs per call. Coverage counts committed positions with reused $q_1$ tree states. These are equal-weight task means.

\begin{table*}[!t]
\centering
\begingroup
\small
\setlength{\tabcolsep}{1.5pt}
\renewcommand{\arraystretch}{1.1}
\begin{tabular*}{\textwidth}{@{}p{0.155\textwidth}p{0.145\textwidth}
@{\extracolsep{\fill}}*{5}{>{\centering\arraybackslash}p{0.070\textwidth}
>{\centering\arraybackslash}p{0.052\textwidth}}@{}}
\toprule
\multirow{2}{*}{Model} & \multirow{2}{*}{Configuration}
& \multicolumn{2}{c}{HumanEval+} & \multicolumn{2}{c}{MBPP+}
& \multicolumn{2}{c}{MATH500} & \multicolumn{2}{c}{BBH}
& \multicolumn{2}{c}{GSM8K} \\
\cmidrule(lr){3-4}\cmidrule(lr){5-6}\cmidrule(lr){7-8}
\cmidrule(lr){9-10}\cmidrule(lr){11-12}
& & Speedup & $\gamma$ & Speedup & $\gamma$ & Speedup & $\gamma$
& Speedup & $\gamma$ & Speedup & $\gamma$ \\
\midrule
\multirow{2}{=}{Ouro-1.4B} & \shortstack{w/o $q_1$\\early logits} & $4.96\!\times$ & 6.36 & $3.75\!\times$ & 4.38 & $2.75\!\times$ & 4.29 & $3.32\!\times$ & 5.58 & $2.80\!\times$ & 3.76 \\
 & Full & $4.99\!\times$ & 7.05 & $4.23\!\times$ & 5.19 & $3.09\!\times$ & 5.22 & $3.59\!\times$ & 6.58 & $2.91\!\times$ & 4.20 \\
\midrule
\multirow{2}{=}{Ouro-2.6B} & \shortstack{w/o $q_1$\\early logits} & $5.46\!\times$ & 6.27 & $4.41\!\times$ & 4.19 & $2.70\!\times$ & 3.95 & $4.14\!\times$ & 5.80 & $4.53\!\times$ & 5.08 \\
 & Full & $6.18\!\times$ & 7.22 & $4.73\!\times$ & 4.63 & $3.10\!\times$ & 4.53 & $4.72\!\times$ & 6.88 & $5.14\!\times$ & 5.91 \\
\midrule
\multirow{2}{=}{\shortstack[l]{Ouro-1.4B-\\Thinking}} & \shortstack{w/o $q_1$\\early logits} & $3.35\!\times$ & 3.63 & $3.51\!\times$ & 3.63 & $2.94\!\times$ & 3.81 & $2.82\!\times$ & 3.31 & $3.40\!\times$ & 3.70 \\
 & Full & $3.49\!\times$ & 3.96 & $3.54\!\times$ & 3.95 & $3.18\!\times$ & 4.31 & $2.79\!\times$ & 3.34 & $3.65\!\times$ & 4.12 \\
\midrule
\multirow{2}{=}{\shortstack[l]{Ouro-2.6B-\\Thinking}} & \shortstack{w/o $q_1$\\early logits} & $3.40\!\times$ & 3.88 & $3.32\!\times$ & 3.85 & $2.73\!\times$ & 4.08 & $2.87\!\times$ & 3.66 & $3.44\!\times$ & 3.96 \\
 & Full & $3.51\!\times$ & 4.25 & $3.82\!\times$ & 4.24 & $3.24\!\times$ & 4.62 & $2.90\!\times$ & 3.70 & $3.84\!\times$ & 4.44 \\
\midrule
\multirow{2}{=}{huginn-0125} & \shortstack{w/o $q_1$\\early logits} & $4.74\!\times$ & 5.70 & $3.88\!\times$ & 4.46 & $4.58\!\times$ & 6.10 & $4.14\!\times$ & 5.40 & $3.83\!\times$ & 4.55 \\
 & Full & $5.22\!\times$ & 6.63 & $4.34\!\times$ & 5.25 & $5.71\!\times$ & 8.81 & $4.48\!\times$ & 6.18 & $4.29\!\times$ & 5.43 \\
\midrule
\multirow{2}{=}{\shortstack[l]{Raven-\\Llama-3.2}} & \shortstack{w/o $q_1$\\early logits} & $4.47\!\times$ & 6.42 & $4.01\!\times$ & 5.31 & $4.31\!\times$ & 5.98 & $4.51\!\times$ & 6.25 & $3.43\!\times$ & 4.50 \\
 & Full & $5.49\!\times$ & 8.44 & $4.80\!\times$ & 6.92 & $5.38\!\times$ & 8.28 & $4.97\!\times$ & 7.54 & $3.80\!\times$ & 5.29 \\
\midrule
\multirow{2}{=}{\shortstack[l]{Recurrent-\\TinyLlama-3T}} & \shortstack{w/o $q_1$\\early logits} & $4.93\!\times$ & 6.29 & $3.97\!\times$ & 4.86 & $4.78\!\times$ & 6.09 & $4.88\!\times$ & 6.43 & $4.17\!\times$ & 5.17 \\
 & Full & $5.47\!\times$ & 7.71 & $4.57\!\times$ & 5.97 & $6.00\!\times$ & 8.83 & $5.71\!\times$ & 8.23 & $4.75\!\times$ & 6.56 \\
\midrule
\multirow{2}{=}{\shortstack[l]{Recurrent-\\OLMo-2}} & \shortstack{w/o $q_1$\\early logits} & $6.38\!\times$ & 11.92 & $6.69\!\times$ & 12.20 & $5.30\!\times$ & 10.40 & $5.14\!\times$ & 10.60 & $5.66\!\times$ & 10.32 \\
 & Full & $7.24\!\times$ & 16.44 & $8.26\!\times$ & 19.45 & $6.11\!\times$ & 15.16 & $6.01\!\times$ & 15.39 & $6.36\!\times$ & 13.58 \\
\bottomrule
\end{tabular*}
\endgroup
\caption{Complete per-task $q_1$ component comparisons. Each task reports speedup over SGLang AR and stage-normalized $\gamma$. Full uses the same single-component reference as the component summaries. The control removes $q_1$-logit updates while retaining token-pattern memory and tree verification.}
\label{tab:q1-update-ablation-full}
\end{table*}

\subsection{Draft-Quality Motivation}
\label{app:motivation}

Figure~\ref{fig:foresight-motivation}(a) uses shared prompts with order-reversed repetitions and fixed 8-node, depth-6 trees. At each observed context, two candidate policies are compared against the same final prefix before verification. Unobserved suffixes are censored, and the known root is excluded. Coverage counts are pooled within each task before averaging the three tasks equally. Panel (b) sums ordinary cycle counts over matched full-system benchmark runs at equal final output lengths. Raven-Llama-3.2 and OLMo use four and eight recurrent stages, respectively.

After censoring OLMo observations at or beyond the onset of an exact periodic run lasting at least 256 tokens, five-token prefix coverage rises from 9.8\% without $q_1$ updates to 25.2\% with them.

\section{Supplementary Experiments}
\label{app:supplementary-experiments}

\subsection{Model Structures}

Raven-Llama-3.2, Recurrent-TinyLlama-3T, and Recurrent-OLMo-2 use a four-layer prelude and a four-layer coda around a shared recurrent core. All three have hidden width 2,048 and retain $S=32$ recurrent iterations. Raven-Llama and OLMo repeat a six-layer core, while TinyLlama repeats an eight-layer core. Raven-Llama and TinyLlama use grouped-query attention with eight and four KV heads, respectively. OLMo uses 16-head multi-head attention.

\subsection{\texorpdfstring{Proposal Depth and Pipeline Granularity}{Proposal Depth and Pipeline Granularity}}
\label{app:pipeline-granularity}

We compare LoopSpec and our method on Raven-Llama-3.2 and Recurrent-OLMo-2 across HumanEval+, MBPP+, MATH500, BBH, and GSM8K. With $S=32$ fixed, $q_1=K\in\{2,4,8,16\}$ gives $P=S/K\in\{16,8,4,2\}$ pipeline stages. Both methods use the same $q_1$ depth and $K$ at each comparison, and $q_2$ remains at iteration 16 when distinct from $q_1$. Our tree budget stays at eight nodes and depth six. Generation uses $T=0$ with the main generation and context limits. Table~\ref{tab:pipeline-granularity} gives equally weighted five-benchmark means.

\begin{table*}[!t]
\centering
\begingroup
\small
\setlength{\tabcolsep}{3pt}
\renewcommand{\arraystretch}{1.12}
\begin{tabular*}{\textwidth}{@{}lrr@{\extracolsep{\fill}}rrrrrr@{}}
\toprule
\multirow{2}{*}{Model} & \multirow{2}{*}{\shortstack{$q_1$ depth\\$=K$}} & \multirow{2}{*}{$P$} & \multicolumn{2}{c}{Speedup over AR} & \multicolumn{2}{c}{$\gamma$} & \multirow{2}{*}{Gain} & \multirow{2}{*}{\shortstack{Cycles/token\\reduction}} \\
\cmidrule(lr){4-5}\cmidrule(lr){6-7}
 & & & LoopSpec & LoopAhead & LoopSpec & LoopAhead & & \\
\midrule
\multirow{4}{*}{\shortstack[l]{Raven-\\Llama-3.2}} & 2 & 16 & $5.43\!\times$ & $5.59\!\times$ & 11.74 & 18.06 & $1.02\!\times$ & 32.1\% \\
 & 4 & 8 & $4.85\!\times$ & $\mathbf{6.50}\!\times$ & 7.32 & 13.55 & $1.33\!\times$ & 43.1\% \\
 & $8^\dagger$ & 4 & $3.14\!\times$ & $4.89\!\times$ & 3.91 & 7.29 & $1.56\!\times$ & 45.0\% \\
 & 16 & 2 & $1.78\!\times$ & $3.09\!\times$ & 1.98 & 3.90 & $1.73\!\times$ & 46.7\% \\
\midrule
\multirow{4}{*}{\shortstack[l]{Recurrent-\\OLMo-2}} & 2 & 16 & $5.05\!\times$ & $5.68\!\times$ & 12.31 & 20.30 & $1.11\!\times$ & 36.4\% \\
 & $4^\dagger$ & 8 & $4.46\!\times$ & $\mathbf{6.80}\!\times$ & 7.51 & 16.00 & $1.52\!\times$ & 52.5\% \\
 & 8 & 4 & $2.99\!\times$ & $5.29\!\times$ & 3.94 & 8.93 & $1.77\!\times$ & 54.2\% \\
 & 16 & 2 & $1.74\!\times$ & $3.45\!\times$ & 1.99 & 4.57 & $1.99\!\times$ & 54.8\% \\
\bottomrule
\end{tabular*}
\endgroup
\caption{Joint variation of $q_1$ depth and pipeline granularity on five benchmarks at $T=0$. Both methods share $K$ and $P$ within each row. $q_1=K$, with $q_2$ at loop 16 when distinct. AR speedups average task-specific speedups. Gain averages paired LoopAhead/LoopSpec ratios at the same $K$. $\dagger$ marks component-ablation references. Bold marks each model's peak mean AR speedup.}
\label{tab:pipeline-granularity}
\end{table*}

\begin{figure}[!htb]
\centering
\includegraphics[width=\columnwidth]{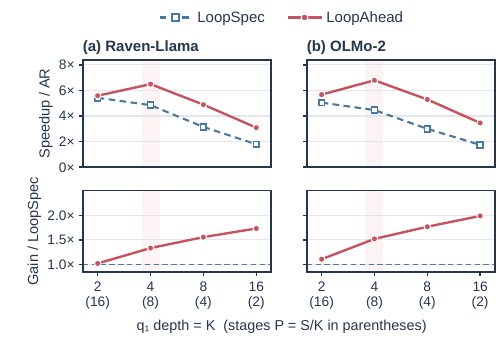}
\caption{Joint variation of $q_1$ depth and pipeline granularity at $S=32$ and $T=0$. Top: mean speedup over SGLang AR. Bottom: mean paired speedup over LoopSpec at the same $K$. Shading marks $q_1=K=4$, with $P=32/K$.}
\label{fig:pipeline-granularity}
\end{figure}

\begin{table*}[t]
\centering
\begingroup
\small
\setlength{\tabcolsep}{1.5pt}
\renewcommand{\arraystretch}{1.2}
\begin{tabular*}{\textwidth}{@{}ll@{\extracolsep{\fill}}*{6}{cc}@{}}
\toprule
\multirow{2}{*}{\textbf{Model}} & \multirow{2}{*}{\textbf{Method}}
& \multicolumn{2}{c}{\textbf{HumanEval+}} & \multicolumn{2}{c}{\textbf{MBPP+}}
& \multicolumn{2}{c}{\textbf{MATH500}} & \multicolumn{2}{c}{\textbf{BBH}}
& \multicolumn{2}{c}{\textbf{GSM8K}} & \multicolumn{2}{c}{\textbf{Average}} \\
\cmidrule(lr){3-4}\cmidrule(lr){5-6}\cmidrule(lr){7-8}
\cmidrule(lr){9-10}\cmidrule(lr){11-12}\cmidrule(lr){13-14}
& & Speedup & $\gamma$ & Speedup & $\gamma$ & Speedup & $\gamma$
& Speedup & $\gamma$ & Speedup & $\gamma$ & Speedup & $\gamma$ \\
\midrule
 & LoopSpec & $2.55\!\times$ & 3.35 & $2.53\!\times$ & 3.15 & $1.91\!\times$ & 3.13 & $1.80\!\times$ & 2.93 & $2.10\!\times$ & 3.04 & $2.18\!\times$ & 3.12 \\
\multirow{-2}{*}{Ouro-1.4B} & \textbf{Ours} & $\mathbf{3.20}\!\times$ & \textbf{4.51} & $\mathbf{2.90}\!\times$ & \textbf{3.87} & $\mathbf{2.30}\!\times$ & \textbf{3.93} & $\mathbf{2.33}\!\times$ & \textbf{4.02} & $\mathbf{2.43}\!\times$ & \textbf{3.57} & $\mathbf{2.63}\!\times$ & \textbf{3.98} \\
\midrule
 & LoopSpec & $2.91\!\times$ & 3.45 & $2.76\!\times$ & 3.22 & $2.17\!\times$ & 3.35 & $2.15\!\times$ & 3.25 & $2.66\!\times$ & 3.41 & $2.53\!\times$ & 3.34 \\
\multirow{-2}{*}{Ouro-2.6B} & \textbf{Ours} & $\mathbf{4.26}\!\times$ & \textbf{4.84} & $\mathbf{3.51}\!\times$ & \textbf{3.68} & $\mathbf{2.88}\!\times$ & \textbf{4.23} & $\mathbf{3.19}\!\times$ & \textbf{4.82} & $\mathbf{3.46}\!\times$ & \textbf{4.23} & $\mathbf{3.46}\!\times$ & \textbf{4.36} \\
\midrule
 & LoopSpec & $2.57\!\times$ & 3.07 & $2.65\!\times$ & 3.05 & $2.28\!\times$ & 3.19 & $2.12\!\times$ & 2.68 & $2.58\!\times$ & 3.09 & $2.44\!\times$ & 3.02 \\
\multirow{-2}{*}{\shortstack[l]{Ouro-1.4B-\\Thinking}} & \textbf{Ours} & $\mathbf{3.04}\!\times$ & \textbf{3.49} & $\mathbf{3.11}\!\times$ & \textbf{3.46} & $\mathbf{2.80}\!\times$ & \textbf{3.85} & $\mathbf{2.37}\!\times$ & \textbf{2.84} & $\mathbf{3.13}\!\times$ & \textbf{3.61} & $\mathbf{2.89}\!\times$ & \textbf{3.45} \\
\midrule
 & LoopSpec & $2.63\!\times$ & 3.42 & $2.74\!\times$ & 3.40 & $2.25\!\times$ & 3.48 & $2.26\!\times$ & 3.13 & $2.64\!\times$ & 3.45 & $2.50\!\times$ & 3.38 \\
\multirow{-2}{*}{\shortstack[l]{Ouro-2.6B-\\Thinking}} & \textbf{Ours} & $\mathbf{3.32}\!\times$ & \textbf{3.95} & $\mathbf{3.47}\!\times$ & \textbf{3.96} & $\mathbf{2.86}\!\times$ & \textbf{4.18} & $\mathbf{2.75}\!\times$ & \textbf{3.52} & $\mathbf{3.48}\!\times$ & \textbf{4.18} & $\mathbf{3.18}\!\times$ & \textbf{3.96} \\
\midrule
 & LoopSpec & $3.18\!\times$ & 3.63 & $3.11\!\times$ & 3.51 & $2.91\!\times$ & 3.49 & $2.91\!\times$ & 3.47 & $2.99\!\times$ & 3.41 & $3.02\!\times$ & 3.50 \\
\multirow{-2}{*}{huginn-0125} & \textbf{Ours} & $\mathbf{3.78}\!\times$ & \textbf{4.72} & $\mathbf{3.54}\!\times$ & \textbf{4.27} & $\mathbf{3.37}\!\times$ & \textbf{4.54} & $\mathbf{3.36}\!\times$ & \textbf{4.46} & $\mathbf{3.23}\!\times$ & \textbf{3.94} & $\mathbf{3.46}\!\times$ & \textbf{4.39} \\
\midrule
 & LoopSpec & $3.03\!\times$ & 3.86 & $3.04\!\times$ & 3.86 & $3.01\!\times$ & 3.88 & $3.07\!\times$ & 3.88 & $3.08\!\times$ & 3.83 & $3.05\!\times$ & 3.86 \\
\multirow{-2}{*}{\shortstack[l]{Raven-\\Llama-3.2}} & \textbf{Ours} & $\mathbf{3.22}\!\times$ & \textbf{4.72} & $\mathbf{3.12}\!\times$ & \textbf{4.46} & $\mathbf{3.27}\!\times$ & \textbf{4.96} & $\mathbf{3.44}\!\times$ & \textbf{5.19} & $\mathbf{3.12}\!\times$ & \textbf{4.37} & $\mathbf{3.23}\!\times$ & \textbf{4.74} \\
\midrule
 & LoopSpec & $3.35\!\times$ & 3.88 & $3.33\!\times$ & 3.84 & $3.26\!\times$ & 3.89 & $3.31\!\times$ & 3.88 & $3.33\!\times$ & 3.85 & $3.32\!\times$ & 3.87 \\
\multirow{-2}{*}{\shortstack[l]{Recurrent-\\TinyLlama-3T}} & \textbf{Ours} & $\mathbf{3.60}\!\times$ & \textbf{4.90} & $\mathbf{3.56}\!\times$ & \textbf{4.75} & $\mathbf{3.51}\!\times$ & \textbf{4.96} & $\mathbf{3.80}\!\times$ & \textbf{5.39} & $\mathbf{3.38}\!\times$ & \textbf{4.52} & $\mathbf{3.57}\!\times$ & \textbf{4.90} \\
\midrule
 & LoopSpec & $4.07\!\times$ & 6.58 & $4.12\!\times$ & 6.49 & $3.72\!\times$ & 6.71 & $3.57\!\times$ & 6.53 & $4.05\!\times$ & 6.74 & $3.91\!\times$ & 6.61 \\
\multirow{-2}{*}{\shortstack[l]{Recurrent-\\OLMo-2}} & \textbf{Ours} & $\mathbf{4.52}\!\times$ & \textbf{7.39} & $\mathbf{4.67}\!\times$ & \textbf{7.37} & $\mathbf{4.22}\!\times$ & \textbf{8.10} & $\mathbf{4.06}\!\times$ & \textbf{7.96} & $\mathbf{4.34}\!\times$ & \textbf{7.46} & $\mathbf{4.36}\!\times$ & \textbf{7.66} \\
\bottomrule
\end{tabular*}
\endgroup
\caption{Full temperature-$1$ results (top-$p=1$, no top-$k$ truncation, seed 1234). Speedups are relative to SGLang AR and $\gamma$ is stage-normalized commitment. Average equally weights the five displayed tasks. Bold marks the higher metric within each model pair.}
\label{tab:temperature-one-full}
\end{table*}

Figure~\ref{fig:pipeline-granularity} separates AR-normalized speedup from the gain over LoopSpec at each matched $K$. Our mean speedup over matched LoopSpec grows from $1.02$--$1.11\times$ at $q_1=K=2$ to $1.73$--$1.99\times$ at $q_1=K=16$. AR-normalized speedup peaks at $q_1=K=4$ ($P=8$) for both models, with $1.33$--$1.52\times$ speedups over LoopSpec and $43.1$--$52.5\%$ fewer cycles per token at the same $K$. Increasing $K$ reduces the number of stages but increases work per stage. Because it also changes the $q_1$ readout depth, this sweep measures the joint effect of drafting depth and pipeline granularity.

\subsection{Complete Results at \texorpdfstring{$T=1$}{T=1}}
\label{app:sampling-results}

Table~\ref{tab:temperature-one-full} reports all eight checkpoints on the five benchmarks at $T=1$. Generation limits, normal EOS, and task stop strings follow the main experiment. The sampling comparison retains $q_1/q_2=4/16$ for OLMo.

\FloatBarrier
\twocolumn[{
  \subsection{Controlled Comparison with LoopSpec}
  \label{app:controlled-comparison}
  \begin{appendixwide}
    \begin{table*}[!t]
\centering
\begingroup
\definecolor{LoopAheadBlue}{rgb}{0.88,0.94,1.0}
\fontsize{8.5}{10}\selectfont
\setlength{\tabcolsep}{1.2pt}
\renewcommand{\arraystretch}{1.12}
\begin{tabularx}{\textwidth}{ll*{6}{r>{\raggedleft\arraybackslash}X}}
\toprule
\multirow{2}{*}{\textbf{Model}} & \multirow{2}{*}{\textbf{Method}}
& \multicolumn{2}{c}{\textbf{HumanEval+}}
& \multicolumn{2}{c}{\textbf{MBPP+}} & \multicolumn{2}{c}{\textbf{MATH500}}
& \multicolumn{2}{c}{\textbf{BBH}} & \multicolumn{2}{c}{\textbf{GSM8K}}
& \multicolumn{2}{c}{\textbf{Average}} \\
\cmidrule(lr){3-4}\cmidrule(lr){5-6}\cmidrule(lr){7-8}
\cmidrule(lr){9-10}\cmidrule(lr){11-12}\cmidrule(lr){13-14}
 & & \multicolumn{1}{c}{Speedup} & \multicolumn{1}{c}{$\gamma$} & \multicolumn{1}{c}{Speedup} & \multicolumn{1}{c}{$\gamma$} & \multicolumn{1}{c}{Speedup} & \multicolumn{1}{c}{$\gamma$} & \multicolumn{1}{c}{Speedup} & \multicolumn{1}{c}{$\gamma$} & \multicolumn{1}{c}{Speedup} & \multicolumn{1}{c}{$\gamma$} & \multicolumn{1}{c}{Speedup} & \multicolumn{1}{c}{$\gamma$} \\
\midrule
 & LoopSpec & $2.97\!\times$ & 3.81 & $2.90\!\times$ & 3.59 & $2.13\!\times$ & 3.47 & $2.20\!\times$ & 3.58 & $2.31\!\times$ & 3.29 & $2.50\!\times$ & 3.55 \\
\rowcolor{LoopAheadBlue}
\cellcolor{white}\multirow{-2}{*}{Ouro-1.4B} & \textbf{LoopAhead} & $\mathbf{4.99}\!\times$ & \textbf{7.05} & $\mathbf{4.23}\!\times$ & \textbf{5.19} & $\mathbf{3.09}\!\times$ & \textbf{5.22} & $\mathbf{3.59}\!\times$ & \textbf{6.58} & $\mathbf{2.91}\!\times$ & \textbf{4.20} & $\mathbf{3.76}\!\times$ & \textbf{5.65} \\
\midrule
 & LoopSpec & $3.23\!\times$ & 3.85 & $3.29\!\times$ & 3.66 & $2.23\!\times$ & 3.47 & $2.50\!\times$ & 3.68 & $2.94\!\times$ & 3.66 & $2.84\!\times$ & 3.66 \\
\rowcolor{LoopAheadBlue}
\cellcolor{white}\multirow{-2}{*}{Ouro-2.6B} & \textbf{LoopAhead} & $\mathbf{6.18}\!\times$ & \textbf{7.22} & $\mathbf{4.73}\!\times$ & \textbf{4.63} & $\mathbf{3.10}\!\times$ & \textbf{4.53} & $\mathbf{4.72}\!\times$ & \textbf{6.88} & $\mathbf{5.14}\!\times$ & \textbf{5.91} & $\mathbf{4.77}\!\times$ & \textbf{5.83} \\
\midrule
 & LoopSpec & $3.38\!\times$ & 3.81 & $3.24\!\times$ & 3.60 & $3.15\!\times$ & 3.78 & $3.09\!\times$ & 3.66 & $3.13\!\times$ & 3.57 & $3.20\!\times$ & 3.68 \\
\rowcolor{LoopAheadBlue}
\cellcolor{white}\multirow{-2}{*}{huginn-0125} & \textbf{LoopAhead} & $\mathbf{5.22}\!\times$ & \textbf{6.63} & $\mathbf{4.34}\!\times$ & \textbf{5.25} & $\mathbf{5.71}\!\times$ & \textbf{8.81} & $\mathbf{4.48}\!\times$ & \textbf{6.18} & $\mathbf{4.29}\!\times$ & \textbf{5.43} & $\mathbf{4.81}\!\times$ & \textbf{6.46} \\
\midrule
 & LoopSpec & $3.20\!\times$ & 3.95 & $3.18\!\times$ & 3.89 & $3.16\!\times$ & 3.94 & $3.08\!\times$ & 3.92 & $3.09\!\times$ & 3.83 & $3.14\!\times$ & 3.91 \\
\rowcolor{LoopAheadBlue}
\cellcolor{white}\multirow{-2}{*}{\shortstack[l]{Raven-\\Llama-3.2}} & \textbf{LoopAhead} & $\mathbf{5.49}\!\times$ & \textbf{8.44} & $\mathbf{4.80}\!\times$ & \textbf{6.92} & $\mathbf{5.38}\!\times$ & \textbf{8.28} & $\mathbf{4.97}\!\times$ & \textbf{7.54} & $\mathbf{3.80}\!\times$ & \textbf{5.29} & $\mathbf{4.89}\!\times$ & \textbf{7.29} \\
\midrule
 & LoopSpec & $3.39\!\times$ & 3.94 & $3.36\!\times$ & 3.85 & $3.36\!\times$ & 3.94 & $3.34\!\times$ & 3.93 & $3.38\!\times$ & 3.87 & $3.36\!\times$ & 3.91 \\
\rowcolor{LoopAheadBlue}
\cellcolor{white}\multirow{-2}{*}{\shortstack[l]{Recurrent-\\TinyLlama-3T}} & \textbf{LoopAhead} & $\mathbf{5.47}\!\times$ & \textbf{7.71} & $\mathbf{4.57}\!\times$ & \textbf{5.97} & $\mathbf{6.00}\!\times$ & \textbf{8.83} & $\mathbf{5.71}\!\times$ & \textbf{8.23} & $\mathbf{4.75}\!\times$ & \textbf{6.56} & $\mathbf{5.30}\!\times$ & \textbf{7.46} \\
\midrule
 & LoopSpec & $3.07\!\times$ & 3.97 & $3.09\!\times$ & 3.97 & $2.90\!\times$ & 3.94 & $2.86\!\times$ & 3.91 & $3.05\!\times$ & 3.90 & $2.99\!\times$ & 3.94 \\
\rowcolor{LoopAheadBlue}
\cellcolor{white}\multirow{-2}{*}{\shortstack[l]{Recurrent-\\OLMo-2}} & \textbf{LoopAhead} & $\mathbf{5.81}\!\times$ & \textbf{9.64} & $\mathbf{6.65}\!\times$ & \textbf{11.78} & $\mathbf{5.10}\!\times$ & \textbf{8.90} & $\mathbf{4.63}\!\times$ & \textbf{7.88} & $\mathbf{4.28}\!\times$ & \textbf{6.45} & $\mathbf{5.29}\!\times$ & \textbf{8.93} \\
\bottomrule
\end{tabularx}
\endgroup
\caption{Controlled greedy comparison with matched recurrent and proposal schedules. Speedups are relative to SGLang AR; $\gamma$ is stage-normalized commitment. Average equally weights the five unrounded task metrics.}
\label{tab:controlled-comparison}
\end{table*}

  \end{appendixwide}
  \vspace{10pt}
}]

Both methods retain full target recurrence and share tokenized inputs (Appendix~\ref{app:execution-protocol}). All models use four recurrent stages, with $q_1$ and $q_2$ read at $S/4$ and $S/2$ loops ($K=S/4$). LoopAhead uses 8-node, depth-6 trees. Greedy generation respects EOS and task stops, with the main length limits.

LoopAhead is faster in all 30 pairs, with model-wise mean gains of $1.50$--$1.77\times$ from equally weighted task-wise ratios. Table~\ref{tab:main-performance} reports the comparison with tuned LoopSpec.

\FloatBarrier

\FloatBarrier
\twocolumn[{
\section{Sanity Checks}
\label{app:sanity-checks}

\subsection{Task Quality under Greedy Decoding}
\label{app:sanity-task-quality}

\begin{appendixwide}
\begin{table*}[t]
\centering
\begingroup
\small
\setlength{\tabcolsep}{6pt}
\renewcommand{\arraystretch}{1.0}
\begin{tabular*}{\textwidth}{@{\extracolsep{\fill}}lrrrrrr@{}}
\toprule
 Model & \multicolumn{2}{c}{HumanEval+} & \multicolumn{2}{c}{MBPP+} & \multicolumn{2}{c}{GSM8K} \\
\cmidrule(lr){2-3}\cmidrule(lr){4-5}\cmidrule(lr){6-7}
& AR & LoopAhead & AR & LoopAhead & AR & LoopAhead \\
\midrule
 Ouro-1.4B & 45.73 & 46.95 & 53.13 & 53.13 & 79.69 & 75.78 \\
Ouro-2.6B & 51.83 & 51.83 & 53.91 & 53.91 & 56.25 & 57.81 \\
Huginn & 14.63 & 14.63 & 32.81 & 32.81 & 40.63 & 40.63 \\
Raven-Llama & 10.37 & 10.37 & 26.56 & 26.56 & 53.91 & 53.91 \\
TinyLlama & 12.80 & 12.80 & 24.22 & 24.22 & 46.88 & 46.88 \\
OLMo-2 & 10.37 & 10.37 & 10.94 & 10.94 & 42.19 & 42.19 \\
\bottomrule
\end{tabular*}
\endgroup
\caption{Zero-shot greedy task scores (\%) for AR and LoopAhead on matched inputs. HumanEval+ and MBPP+ use EvalPlus pass@1, and GSM8K uses exact-match accuracy.}
\label{tab:sanity-task-quality}
\end{table*}

\end{appendixwide}
\vspace{10pt}
}]

We compare task quality with SGLang autoregressive decoding (AR) under zero-shot greedy generation. Table~\ref{tab:sanity-task-quality} reports EvalPlus pass@1 on HumanEval+ and MBPP+, using the combined base and additional tests, and exact-match accuracy on GSM8K. These runs use generation/context limits of 1,024/2,048 tokens. Generation respects normal EOS and task stop strings.

\paragraph{Results.}
AR and LoopAhead have identical per-input task verdicts for Huginn, Raven-Llama, TinyLlama, and OLMo-2 across all three benchmarks.

\subsection{Small-Vocabulary Sampling Tests}
\label{app:sanity-sampling}

We test the production sampling decision path using eight small-vocabulary checks with 32,768 samples per check. These cover single-step rejection sampling, residual correction, zero-probability support, top-$k$/top-$p$ filtering, and two-token conditional joint distributions. All goodness-of-fit tests against the analytic target distributions pass after Holm correction. Direct comparisons with independent AR sampling also yield no rejected two-sample tests. The minimum Holm-adjusted $p$-value is $0.527$, and the maximum empirical total variation distance is $0.007111$.

\subsection{Token Agreement and Output Length}
\label{app:sanity-token-agreement}

LoopSpec and LoopAhead produce identical tokens and output lengths for Huginn, Raven-Llama, TinyLlama, and OLMo-2 on these three benchmarks. Raven-Llama and TinyLlama also match native AR token for token and receive identical per-input task verdicts across all three benchmarks. Earlier native-AR outputs on the same MBPP+ inputs likewise match OLMo-2's tokens and verdicts.

\begin{appendixcolumn}
\begin{table}[htbp]
\centering
\begingroup
\small
\setlength{\tabcolsep}{3pt}
\renewcommand{\arraystretch}{0.9}
\begin{tabular*}{\columnwidth}{@{\extracolsep{\fill}}llrr@{}}
\toprule
 Model & Benchmark & LoopSpec & LoopAhead \\
\midrule
 Ouro-1.4B & HumanEval+ & 202.58 & 212.09 \\
 & MBPP+ & 96.97 & 89.70 \\
 & GSM8K & 589.58 & 610.06 \\
\midrule
 Ouro-2.6B & HumanEval+ & 231.11 & 232.16 \\
 & MBPP+ & 85.61 & 73.34 \\
 & GSM8K & 176.53 & 184.09 \\
\midrule
 Huginn & HumanEval+ & 252.93 & 252.93 \\
 & MBPP+ & 106.75 & 106.75 \\
 & GSM8K & 131.13 & 131.13 \\
\midrule
 Raven-Llama & HumanEval+ & 373.48 & 373.48 \\
 & MBPP+ & 156.24 & 156.24 \\
 & GSM8K & 93.63 & 93.63 \\
\midrule
 TinyLlama & HumanEval+ & 297.35 & 297.35 \\
 & MBPP+ & 103.81 & 103.81 \\
 & GSM8K & 130.95 & 130.95 \\
\midrule
 OLMo-2 & HumanEval+ & 407.98 & 407.98 \\
 & MBPP+ & 367.66 & 367.66 \\
 & GSM8K & 245.60 & 245.60 \\
\bottomrule
\end{tabular*}
\endgroup
\caption{Mean generated lengths (tokens) for LoopSpec and LoopAhead, with normal EOS and task stops.}
\label{tab:sanity-output-length}
\end{table}

\end{appendixcolumn}

\subsection{Numerical Behavior in SGLang}
\label{app:ouro-numerical-drift}

\paragraph{Shape-dependent execution in SGLang.}
In SGLang's BF16 execution, batch shape can affect projection-kernel selection and split-KV attention reductions. Our scheduler changes batch composition while retaining the full target recurrence and causal visibility. The reduction and rounding orders can differ from token-wise execution even when the mathematical operations are unchanged.

\paragraph{Controlled evidence and task quality.}
Earlier controlled Ouro replays isolated this effect under identical Q/K/V and logical prefix KV. Ordinary and block execution used two and one KV splits, respectively, giving an attention-output difference of $5.96\times10^{-8}$. A common split count eliminated this local difference across both physical KV layouts. Separate probes also found batch-width-dependent rounding in the attention-output and MLP-down projections. Such differences can reverse near-tied final-logit rankings and change subsequent greedy contexts. 

\subsection{Paired Component Controls}
\label{app:paired-component-controls}

\begin{table}[!ht]
\centering
\begingroup
\small
\setlength{\tabcolsep}{1.5pt}
\renewcommand{\arraystretch}{1.12}
\begin{tabular*}{\columnwidth}{@{}ccc@{\extracolsep{\fill}}ccc@{}}
\toprule
$H$ & $Q$ & $B$ & Raven-Llama & TinyLlama & OLMo-2 \\
\midrule
0 & 0 & 0 & \shortstack{$1.00\!\times$\\{\scriptsize [1.00, 1.00]}} & \shortstack{$1.00\!\times$\\{\scriptsize [1.00, 1.00]}} & \shortstack{$1.00\!\times$\\{\scriptsize [1.00, 1.00]}} \\
0 & 0 & 1 & \shortstack{$1.00\!\times$\\{\scriptsize [0.95, 1.04]}} & \shortstack{$0.99\!\times$\\{\scriptsize [0.99, 1.00]}} & \shortstack{$1.00\!\times$\\{\scriptsize [1.00, 1.00]}} \\
0 & 1 & 0 & \shortstack{$0.98\!\times$\\{\scriptsize [0.95, 1.01]}} & \shortstack{$0.97\!\times$\\{\scriptsize [0.97, 0.98]}} & \shortstack{$0.94\!\times$\\{\scriptsize [0.94, 0.95]}} \\
0 & 1 & 1 & \shortstack{$1.32\!\times$\\{\scriptsize [1.27, 1.38]}} & \shortstack{$1.32\!\times$\\{\scriptsize [1.31, 1.32]}} & \shortstack{$1.48\!\times$\\{\scriptsize [1.48, 1.49]}} \\
1 & 0 & 0 & \shortstack{$0.99\!\times$\\{\scriptsize [0.97, 1.02]}} & \shortstack{$0.98\!\times$\\{\scriptsize [0.98, 0.99]}} & \shortstack{$0.95\!\times$\\{\scriptsize [0.95, 0.95]}} \\
1 & 0 & 1 & \shortstack{$1.25\!\times$\\{\scriptsize [1.23, 1.28]}} & \shortstack{$1.23\!\times$\\{\scriptsize [1.23, 1.23]}} & \shortstack{$1.27\!\times$\\{\scriptsize [1.26, 1.27]}} \\
1 & 1 & 0 & \shortstack{$0.98\!\times$\\{\scriptsize [0.96, 1.00]}} & \shortstack{$0.97\!\times$\\{\scriptsize [0.96, 0.97]}} & \shortstack{$0.94\!\times$\\{\scriptsize [0.94, 0.94]}} \\
1 & 1 & 1 & \shortstack{$1.36\!\times$\\{\scriptsize [1.32, 1.39]}} & \shortstack{$1.35\!\times$\\{\scriptsize [1.33, 1.36]}} & \shortstack{$1.51\!\times$\\{\scriptsize [1.50, 1.51]}} \\
\bottomrule
\end{tabular*}
\endgroup
\caption{Complete component controls. $H$ enables historical continuations, $Q$ enables stored $q_1$ candidates, and $B$ enables multi-token advancement and commitment. Throughput is normalized to $H=Q=B=0$ within each paired run. Values give geometric mean gains and 95\% intervals across three runs.}
\label{tab:component-factorial-full}
\end{table}

We evaluate all eight on/off settings of three components on Raven-Llama, TinyLlama, and OLMo across HumanEval+, MBPP+, MATH500, BBH, and GSM8K. Each benchmark uses 32 fixed prompts selected from the main input set with seed 5678. Target recurrence, proposal depths, and tree shape remain fixed at $S=32$, $q_1/q_2=4/8$, eight nodes, depth six, and reserve two. Generation uses $T=0$, normal EOS, and the main generation/context limits. Each model has three complete paired runs, with all settings in a run sharing tokenized inputs, numerical kernels, and the same physical GPU.

$H$ enables observed historical continuations from the prompt and committed output. Disabling $H$ retains the contextual keys and suffix lookup used by stored predictions. $Q$ enables stored $q_1$ candidates. Disabling $Q$ retains ordinary shallow readout and tree verification. $B$ enables multi-input advancement and multi-token commitment. All eight settings produce identical output tokens within each run.

For a component contrast in run $j$, we equally weight the five benchmark log-throughput ratios to obtain $u_j$. The reported gain is $\exp(\bar u)$, with a 95\% interval of $\exp(\bar u\pm t_{0.975,2}s_u/\sqrt{3})$. Table~\ref{tab:paired-components-3090} enables one component with the other two active. Table~\ref{tab:component-factorial-full} gives the complete matrix normalized to the setting with all three disabled.

Two-sided paired $t$-tests use the three run-level contrasts, with Holm correction across the nine model--component comparisons.

With history enabled, the $q_1$ cache gives $0.98$--$0.99\times$ under token-wise advancement and $1.08$--$1.19\times$ when selected inputs advance together. The corresponding paired interaction ratios are $1.10$--$1.21$, with all three 95\% intervals above one.

\FloatBarrier

\twocolumn[{
  \section{Thinking Checkpoints}
  \label{app:thinking}
  \begin{appendixwide}
    \begin{table*}[!t]
\centering
\begingroup
\small
\setlength{\tabcolsep}{1.5pt}
\renewcommand{\arraystretch}{1.18}
\begin{tabular*}{\textwidth}{@{}ll@{\extracolsep{\fill}}*{6}{cc}@{}}
\toprule
\multirow{2}{*}{\textbf{Model}} & \multirow{2}{*}{\textbf{Method}}
& \multicolumn{2}{c}{\textbf{HumanEval+}}
& \multicolumn{2}{c}{\textbf{MBPP+}} & \multicolumn{2}{c}{\textbf{MATH500}}
& \multicolumn{2}{c}{\textbf{BBH}} & \multicolumn{2}{c}{\textbf{GSM8K}}
& \multicolumn{2}{c}{\textbf{Average}} \\
\cmidrule(lr){3-4}\cmidrule(lr){5-6}\cmidrule(lr){7-8}
\cmidrule(lr){9-10}\cmidrule(lr){11-12}\cmidrule(lr){13-14}
& & Speedup & $\gamma$ & Speedup & $\gamma$ & Speedup & $\gamma$
& Speedup & $\gamma$ & Speedup & $\gamma$ & Speedup & $\gamma$ \\
\midrule
 & SSSD & $0.95\!\times$ & 1.79 & $0.99\!\times$ & 1.75 & $0.80\!\times$ & 1.97 & $0.72\!\times$ & 1.57 & $1.03\!\times$ & 1.96 & $0.90\!\times$ & 1.81 \\
 & NGRAM & $2.07\!\times$ & 2.38 & $2.30\!\times$ & 2.58 & $1.80\!\times$ & 2.39 & $1.85\!\times$ & 2.27 & $2.10\!\times$ & 2.42 & $2.02\!\times$ & 2.41 \\
 & RACER & $2.07\!\times$ & 2.80 & $2.18\!\times$ & 2.82 & $1.75\!\times$ & 2.88 & $1.51\!\times$ & 2.21 & $2.13\!\times$ & 2.92 & $1.93\!\times$ & 2.73 \\
 & LoopSpec & $2.79\!\times$ & 3.25 & $2.84\!\times$ & 3.26 & $2.45\!\times$ & 3.36 & $2.37\!\times$ & 2.97 & $2.85\!\times$ & 3.31 & $2.66\!\times$ & 3.23 \\
\multirow{-5}{*}{\shortstack[l]{Ouro-1.4B-\\Thinking}} & \textbf{Ours} & $\mathbf{3.49}\!\times$ & \textbf{3.96} & $\mathbf{3.54}\!\times$ & \textbf{3.95} & $\mathbf{3.18}\!\times$ & \textbf{4.31} & $\mathbf{2.79}\!\times$ & \textbf{3.34} & $\mathbf{3.65}\!\times$ & \textbf{4.12} & $\mathbf{3.33}\!\times$ & \textbf{3.94} \\
\midrule
 & SSSD & $0.82\!\times$ & 1.77 & $0.87\!\times$ & 1.74 & $0.70\!\times$ & 1.97 & $0.61\!\times$ & 1.53 & $0.91\!\times$ & 1.96 & $0.78\!\times$ & 1.79 \\
 & NGRAM & $1.80\!\times$ & 2.34 & $2.02\!\times$ & 2.54 & $1.59\!\times$ & 2.38 & $1.59\!\times$ & 2.21 & $1.82\!\times$ & 2.36 & $1.76\!\times$ & 2.37 \\
 & RACER & $1.96\!\times$ & 2.82 & $2.03\!\times$ & 2.80 & $1.69\!\times$ & 2.92 & $1.52\!\times$ & 2.34 & $2.06\!\times$ & 2.95 & $1.85\!\times$ & 2.77 \\
 & LoopSpec & $2.65\!\times$ & 3.47 & $2.80\!\times$ & 3.44 & $2.32\!\times$ & 3.54 & $2.31\!\times$ & 3.25 & $2.77\!\times$ & 3.50 & $2.57\!\times$ & 3.44 \\
\multirow{-5}{*}{\shortstack[l]{Ouro-2.6B-\\Thinking}} & \textbf{Ours} & $\mathbf{3.51}\!\times$ & \textbf{4.25} & $\mathbf{3.82}\!\times$ & \textbf{4.24} & $\mathbf{3.24}\!\times$ & \textbf{4.62} & $\mathbf{2.90}\!\times$ & \textbf{3.70} & $\mathbf{3.84}\!\times$ & \textbf{4.44} & $\mathbf{3.46}\!\times$ & \textbf{4.25} \\
\bottomrule
\end{tabular*}
\endgroup
\caption{Greedy Thinking speedups over SGLang AR and stage-normalized commitment $\gamma$. Average equally weights the five displayed tasks. Bold marks the best result per column within each model.}
\label{tab:thinking-performance}
\end{table*}

    \par\vspace{10pt}
    \noindent
    \begin{minipage}[t]{0.48\textwidth}
      \begin{appendixcolumn}
        \begin{table}[t]
\centering
\begingroup
\small
\setlength{\tabcolsep}{1pt}
\renewcommand{\arraystretch}{1.10}
\begin{tabular*}{\columnwidth}{@{}l@{\extracolsep{\fill}}cccc@{}}
\toprule
Model & Full & \shortstack{w/o\\block} & \shortstack{w/o\\$n$-gram}
& \shortstack{w/o $q_1$\\early logits} \\
\midrule
Ouro-1.4B-T & $\mathbf{3.33}\!\times$ & $2.83\!\times$ & $3.24\!\times$ & $3.21\!\times$ \\
Ouro-2.6B-T & $\mathbf{3.46}\!\times$ & $2.85\!\times$ & $3.38\!\times$ & $3.15\!\times$ \\
\bottomrule
\end{tabular*}
\endgroup
\caption{Greedy Thinking component comparisons at $T=0$. Speedups over SGLang AR, averaged equally over five tasks. Full uses the Thinking ablation reference, with separately measured controls. The $q_1$ control removes logit-memory updates, not tree verification.}
\label{tab:thinking-component-ablation}
\end{table}

      \end{appendixcolumn}
    \end{minipage}\hfill
    \begin{minipage}[t]{0.48\textwidth}
      \begin{appendixcolumn}
        \begin{table}[t]
\centering
\begingroup
\small
\setlength{\tabcolsep}{2pt}
\renewcommand{\arraystretch}{1.08}
\begin{tabular*}{\columnwidth}{@{}l@{\extracolsep{\fill}}rrrr@{}}
\toprule
\multirow{2}{*}{Model} & \multicolumn{2}{c}{$T=0$} & \multicolumn{2}{c}{$T=1$} \\
\cmidrule(lr){2-3}\cmidrule(lr){4-5}
& LS & Ours & LS & Ours \\
\midrule
Ouro-1.4B-T & $2.66\!\times$ & $\mathbf{3.33}\!\times$ & $2.44\!\times$ & $\mathbf{2.89}\!\times$ \\
Ouro-2.6B-T & $2.57\!\times$ & $\mathbf{3.46}\!\times$ & $2.50\!\times$ & $\mathbf{3.18}\!\times$ \\
\bottomrule
\end{tabular*}
\endgroup
\caption{Thinking temperature comparison: mean speedups over SGLang AR across five tasks. LS denotes LoopSpec. -T denotes Thinking. Bold marks the faster method at each temperature.}
\label{tab:thinking-temperature-summary}
\end{table}

      \end{appendixcolumn}
    \end{minipage}
  \end{appendixwide}
  \vspace{10pt}
}]

Ouro-1.4B-Thinking and Ouro-2.6B-Thinking use the same five benchmarks and measurement protocol as the main experiment. Both retain all four target loops, checkpoint chat templates, normal EOS, and the 1,024/2,048 generation/context limits.

\subsection{Throughput and Final Commitment}

Table~\ref{tab:thinking-performance} reports mean greedy speedups of $3.33\times$ and $3.46\times$ over SGLang AR. Stage-normalized commitment $\gamma$ increases from 3.23 to 3.94 and from 3.44 to 4.25, respectively. Mean paired gains over LoopSpec are $1.25\times$ and $1.34\times$ under the Thinking reference settings.

The mechanism counters in Table~\ref{tab:q1-final-yield} show a corresponding conversion gap. The two checkpoints have $q_1$/final agreement of 86.4\% and 90.4\% on committed contexts. They locally accept 2.52 and 2.45 tree nodes per call, including the root, but ultimately commit only 0.92 and 1.01 non-root inputs per call. Excluding the root still leaves a gap between local $q_1$ selection and final yield, limiting continuous final commitment.

\subsection{Component Ablations}

In Table~\ref{tab:thinking-component-ablation}, block pipelining provides $1.18\times$ and $1.22\times$ higher mean throughput. Restoring historical $n$-grams provides 2.8\% and 2.3\% gains.

Using the Thinking ablation reference, throughput is 3.9\% and 9.9\% higher than the w/o $q_1$ early logits control. Table~\ref{tab:q1-update-ablation-full} gives the corresponding per-task results under the same reference. Early bonus logits guide continued drafting, while block pipelining advances the resulting multi-token spans.

\subsection{Stochastic Sampling}

At $T=1$, mean speedups over SGLang AR remain $2.89\times$ and $3.18\times$ (Table~\ref{tab:thinking-temperature-summary}), retaining 87\% and 92\% of their greedy throughput. Mean paired gains over LoopSpec are $1.18\times$ and $1.27\times$. Full per-task results and sampling controls appear in Appendix~\ref{app:sampling-results}.

\FloatBarrier
\twocolumn[{
  \section{A800 Results and Hardware Comparison}
  \label{app:a800}
  \begin{appendixwide}
    \begin{table*}[!t]
\centering
\begingroup
\small
\setlength{\tabcolsep}{3pt}
\renewcommand{\arraystretch}{1.12}
\begin{tabular*}{\textwidth}{@{}ll@{\extracolsep{\fill}}rrr@{}}
\toprule
\multirow{2}{*}{Model} & \multirow{2}{*}{Task} & \multicolumn{2}{c}{Speedup over AR} & \multirow{2}{*}{Gain} \\
\cmidrule(lr){3-4}
 & & LoopSpec & LoopAhead & \\
\midrule
\multirow{3}{*}{Ouro-1.4B} & HumanEval & $3.88\!\times$ & $\mathbf{8.00}\!\times$ & $2.06\!\times$ \\
 & MBPP & $4.09\!\times$ & $\mathbf{7.98}\!\times$ & $1.95\!\times$ \\
 & GSM8K & $3.88\!\times$ & $\mathbf{5.70}\!\times$ & $1.47\!\times$ \\
\midrule
\multirow{3}{*}{Ouro-2.6B} & HumanEval & $3.92\!\times$ & $\mathbf{8.63}\!\times$ & $2.20\!\times$ \\
 & MBPP & $4.20\!\times$ & $\mathbf{11.01}\!\times$ & $2.62\!\times$ \\
 & GSM8K & $3.73\!\times$ & $\mathbf{4.56}\!\times$ & $1.22\!\times$ \\
\midrule
\multirow{3}{*}{Huginn} & HumanEval & $3.35\!\times$ & $\mathbf{6.28}\!\times$ & $1.87\!\times$ \\
 & MBPP & $3.47\!\times$ & $\mathbf{7.73}\!\times$ & $2.23\!\times$ \\
 & GSM8K & $3.44\!\times$ & $\mathbf{7.18}\!\times$ & $2.09\!\times$ \\
\midrule
\multirow{3}{*}{Raven-Llama-3.2} & HumanEval & $3.25\!\times$ & $\mathbf{6.19}\!\times$ & $1.91\!\times$ \\
 & MBPP & $3.25\!\times$ & $\mathbf{6.62}\!\times$ & $2.03\!\times$ \\
 & GSM8K & $3.24\!\times$ & $\mathbf{5.73}\!\times$ & $1.77\!\times$ \\
\midrule
\multirow{3}{*}{\shortstack[l]{Recurrent-\\TinyLlama}} & HumanEval & $3.30\!\times$ & $\mathbf{6.14}\!\times$ & $1.86\!\times$ \\
 & MBPP & $3.32\!\times$ & $\mathbf{6.35}\!\times$ & $1.91\!\times$ \\
 & GSM8K & $3.31\!\times$ & $\mathbf{5.86}\!\times$ & $1.77\!\times$ \\
\midrule
\multirow{3}{*}{\shortstack[l]{Recurrent-\\OLMo-2}} & HumanEval & $5.18\!\times$ & $\mathbf{9.00}\!\times$ & $1.74\!\times$ \\
 & MBPP & $5.13\!\times$ & $\mathbf{8.10}\!\times$ & $1.58\!\times$ \\
 & GSM8K & $5.17\!\times$ & $\mathbf{7.65}\!\times$ & $1.48\!\times$ \\
\midrule
\multirow{2}{*}{\shortstack[l]{Ouro-1.4B-\\Thinking}} & MBPP & $3.57\!\times$ & $\mathbf{3.97}\!\times$ & $1.11\!\times$ \\
 & GSM8K & $3.72\!\times$ & $\mathbf{5.02}\!\times$ & $1.35\!\times$ \\
\midrule
\multirow{3}{*}{\shortstack[l]{Ouro-2.6B-\\Thinking}} & HumanEval & $3.73\!\times$ & $\mathbf{4.35}\!\times$ & $1.17\!\times$ \\
 & MBPP & $3.93\!\times$ & $\mathbf{4.76}\!\times$ & $1.21\!\times$ \\
 & GSM8K & $4.09\!\times$ & $\mathbf{5.90}\!\times$ & $1.44\!\times$ \\
\bottomrule
\end{tabular*}
\endgroup
\caption{Full-length results for the 23 model--task pairs. Speedups are relative to SGLang AR, and Gain is LoopAhead/LoopSpec. Ouro-1.4B-Thinking is reported on MBPP and GSM8K.}
\label{tab:a800-full}
\end{table*}

  \end{appendixwide}
  \vspace{10pt}
}]

The A800 evaluation covers all eight checkpoints on GSM8K and MBPP and seven on HumanEval. 

Figure~\ref{fig:hardware-comparison} uses the six non-Thinking checkpoints with all three benchmark families on both GPUs. Both GPUs use $K=1$ for Ouro, $K=8$ for Huginn, Raven-Llama, and TinyLlama, and $K=4$ for OLMo. The A800 bars use the EOS-truncated columns in Table~\ref{tab:a800-eos}. The RTX 3090 summary uses these same 18 model--task cells. Each cell is normalized by its own GPU's AR rate. Each panel averages the six checkpoint ratios for one benchmark. Panel axes start at zero and use task-specific upper limits. The cross-task means below give equal weight to all 18 ratios. 

\begin{samepage}

On these 18 non-Thinking model--task pairs, EOS-truncated mean speedups over AR are $4.03\times$ for LoopSpec and $6.31\times$ for our method on A800. The corresponding RTX 3090 means are $3.39\times$ and $5.16\times$. Direct paired gains over LoopSpec average $1.60\times$ and $1.51\times$, respectively.
\par
\end{samepage}

On the same 18 non-Thinking model--task pairs, full-length A800 results give $3.84\times$ for LoopSpec, $7.15\times$ for our method, and a mean paired gain of $1.88\times$ (Table~\ref{tab:a800-full}).

\subsection{Component Effects on Pipeline Cycles}
\label{app:component-cycle-reduction}
Across six models, historical $n$-grams, $q_1$ updates, and multi-token advancement reduce cycles per generated token by $5.5\%$, $21.8\%$, and $47.1\%$, respectively (Figure~\ref{fig:component-cycle-reduction}).

\twocolumn[{
  \begin{minipage}{\textwidth}
    \centering
    \captionsetup{hypcap=false}

    \includegraphics[width=\textwidth]
      {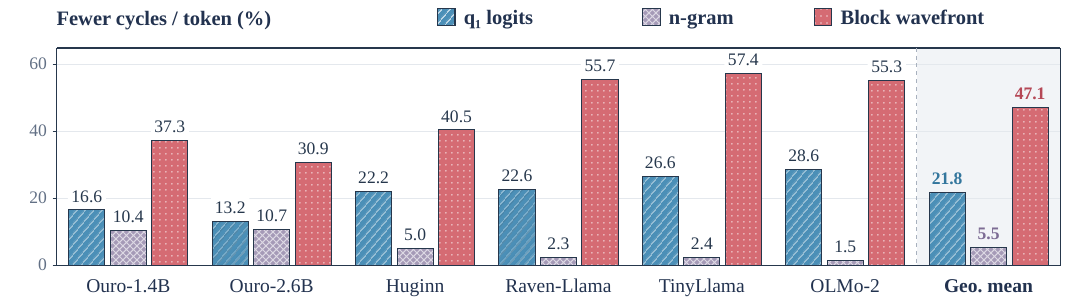}
    \captionof{figure}{Reduction in pipeline cycles per generated
    token when each component is enabled. Reductions use geometric
    means of paired Full/ablated cycle-per-token ratios.
    The aggregate weights models equally.}
    \label{fig:component-cycle-reduction}

    \vspace{6pt}

    \begin{appendixwide}
      \begin{table*}[!t]
\centering
\begingroup
\small
\setlength{\tabcolsep}{3pt}
\renewcommand{\arraystretch}{1.12}
\begin{tabular*}{\textwidth}{@{}ll@{\extracolsep{\fill}}rrr@{}}
\toprule
\multirow{2}{*}{Model} & \multirow{2}{*}{Task} & \multicolumn{2}{c}{Speedup over AR} & \multirow{2}{*}{Gain} \\
\cmidrule(lr){3-4}
 & & LoopSpec & LoopAhead & \\
\midrule
\multirow{3}{*}{Ouro-1.4B} & HumanEval & $4.69\!\times$ & $\mathbf{6.12}\!\times$ & $1.31\!\times$ \\
 & MBPP & $4.79\!\times$ & $\mathbf{5.54}\!\times$ & $1.16\!\times$ \\
 & GSM8K & $4.59\!\times$ & $\mathbf{5.28}\!\times$ & $1.15\!\times$ \\
\midrule
\multirow{3}{*}{Ouro-2.6B} & HumanEval & $4.15\!\times$ & $\mathbf{5.58}\!\times$ & $1.35\!\times$ \\
 & MBPP & $4.84\!\times$ & $\mathbf{5.82}\!\times$ & $1.20\!\times$ \\
 & GSM8K & $4.02\!\times$ & $\mathbf{4.78}\!\times$ & $1.19\!\times$ \\
\midrule
\multirow{3}{*}{Huginn} & HumanEval & $3.44\!\times$ & $\mathbf{5.29}\!\times$ & $1.54\!\times$ \\
 & MBPP & $3.38\!\times$ & $\mathbf{4.38}\!\times$ & $1.29\!\times$ \\
 & GSM8K & $3.31\!\times$ & $\mathbf{4.54}\!\times$ & $1.37\!\times$ \\
\midrule
\multirow{3}{*}{Raven-Llama-3.2} & HumanEval & $3.25\!\times$ & $\mathbf{6.68}\!\times$ & $2.06\!\times$ \\
 & MBPP & $3.26\!\times$ & $\mathbf{7.21}\!\times$ & $2.21\!\times$ \\
 & GSM8K & $3.25\!\times$ & $\mathbf{5.83}\!\times$ & $1.80\!\times$ \\
\midrule
\multirow{3}{*}{\shortstack[l]{Recurrent-\\TinyLlama}} & HumanEval & $3.31\!\times$ & $\mathbf{6.90}\!\times$ & $2.08\!\times$ \\
 & MBPP & $3.33\!\times$ & $\mathbf{7.06}\!\times$ & $2.12\!\times$ \\
 & GSM8K & $3.32\!\times$ & $\mathbf{5.97}\!\times$ & $1.80\!\times$ \\
\midrule
\multirow{3}{*}{\shortstack[l]{Recurrent-\\OLMo-2}} & HumanEval & $5.21\!\times$ & $\mathbf{9.81}\!\times$ & $1.88\!\times$ \\
 & MBPP & $5.16\!\times$ & $\mathbf{9.09}\!\times$ & $1.76\!\times$ \\
 & GSM8K & $5.17\!\times$ & $\mathbf{7.68}\!\times$ & $1.49\!\times$ \\
\midrule
\multirow{2}{*}{\shortstack[l]{Ouro-1.4B-\\Thinking}} & MBPP & $3.60\!\times$ & $\mathbf{3.90}\!\times$ & $1.08\!\times$ \\
 & GSM8K & $3.92\!\times$ & $\mathbf{4.27}\!\times$ & $1.09\!\times$ \\
\midrule
\multirow{3}{*}{\shortstack[l]{Ouro-2.6B-\\Thinking}} & HumanEval & $3.76\!\times$ & $\mathbf{4.33}\!\times$ & $1.15\!\times$ \\
 & MBPP & $4.03\!\times$ & $\mathbf{4.72}\!\times$ & $1.17\!\times$ \\
 & GSM8K & $4.44\!\times$ & $\mathbf{5.19}\!\times$ & $1.17\!\times$ \\
\bottomrule
\end{tabular*}
\endgroup
\caption{EOS-truncated results for the 23 model--task pairs. Speedups are relative to SGLang AR, and Gain is LoopAhead/LoopSpec. The six non-Thinking checkpoints supply the A800 bars in Figure~\ref{fig:hardware-comparison}.}
\label{tab:a800-eos}
\end{table*}

    \end{appendixwide}
  \end{minipage}
  \vspace{4pt}
}]

\clearpage
\section{Baselines and Method Details}
\label{app:verification-semantics}

The baselines cover autoregressive decoding, retrieval- and logit-based drafting, and token-wise recurrent pipelines. Notation follows Section~\ref{sec:preliminaries}, with $P=S/K$, first-stage readout $p_1$, and final target $p=p_P$. Greedy rules use $g_j$ from Eq.~\ref{eq:recurrent-stages}.

\paragraph{SGLang autoregressive decoding (AR).}
SGLang~\citep{zheng2023sglang} provides an efficient serving runtime with KV-cache reuse through RadixAttention. AR denotes non-speculative target decoding in this runtime. For each output token, a looped model executes its native prelude, all $S$ recurrent iterations, and its coda. Greedy decoding then emits $y=g_P(c)$ and extends the context to $c\mathbin{\Vert}y$. Sampling instead draws $y\sim p_P(\cdot\mid c)$. Cached prefix KV avoids recomputing earlier positions, but shared weights do not remove the sequential recurrence within each new position. AR is the throughput reference for the reported speedups.

\paragraph{Simply-Scalable Speculative Decoding (SSSD).}
SSSD~\citep{marzollo2026sssd} constructs drafts through CPU $n$-gram retrieval rather than a shallow neural forward pass. It combines a trie over the prompt and generated output with a suffix-array datastore. For a matched key $a=\kappa_h(c)$ and a candidate continuation $u$, the uncalibrated source score is
\begin{equation}
 w_{\mathcal D}(u\mid a)=
 \frac{\operatorname{count}_{\mathcal D}(a\mathbin{\Vert}u)}
      {\operatorname{count}_{\mathcal D}(a)}.
 \label{eq:baseline-sssd-score}
\end{equation}
Here $\mathcal D$ denotes a retrieval source and the denominator is positive. Source-dependent scaling and adjustments for match length and candidate depth calibrate these scores before candidate trees are merged. A hardware-aware speculation budget accounts for batch size and verification cost. The target verifies the resulting tree. Unlike our method, SSSD does not use intermediate recurrent logits to replenish drafts.

\paragraph{Native $n$-gram speculation (NGRAM).}
NGRAM is SGLang's native $n$-gram speculative decoder.\footnote{\url{https://docs.sglang.io/docs/advanced_features/speculative_decoding\#ngram-speculative-decoding}} It retrieves continuations by matching context suffixes against a token cache. Reusing $\mathcal{H}$ for that method's cached history windows, historical next-token candidates at a tree node satisfy
\begin{equation}
 \mathcal{C}_{\mathrm{hist}}(c)\subseteq
 \bigcup_{h=1}^{\min(n,|c|)}
 \bigl\{x:\kappa_h(c)\mathbin{\Vert}x\in\mathcal{H}\bigr\}.
 \label{eq:baseline-history-candidates}
\end{equation}
Here $\kappa_h$ is the suffix operator from Section~\ref{subsec:method-updates}, and $n$ bounds the matching key length. Recency-based breadth-first or frequency-based expansion selects these matches into a bounded draft tree. The full target verifies its nodes without a separate draft model or neural drafting forward pass. The candidate source is cached token sequences rather than intermediate recurrent logits.

\paragraph{RACER.}
RACER~\citep{zhang2026racer} combines a retrieval tree with a logits tree. An Aho--Corasick automaton indexes observed $n$-grams, uses failure links for suffix matching, and evicts least-recently-used leaves to bound storage. Retrieved continuations are ranked by frequency. To express its copy-logit strategy, let $c=x_{1:t}$ end in the newly proposed token. When the most recent earlier occurrence has a cached readout,
\begin{equation}
 \begin{aligned}
 i^\star &= \max\{i:1\le i<t,\ x_i=x_t\},\\
 \widetilde z_P(c) &= z_P(x_{1:i^\star}).
 \end{aligned}
 \label{eq:baseline-racer-copy-logit}
\end{equation}
The surrogate supplies $\operatorname{Top}_k(\widetilde z_P(c))$ candidates, not a verified readout for $c$. Retrieval receives priority within the draft budget, and remaining slots support logits-tree expansion. Prefix merging produces one tree for target verification. Verified sequences and newly computed logits refresh the two candidate sources.

\paragraph{LoopSpec.}
LoopSpec~\citep{cho2026loopspec} starts a child lane from an early $q_1$ prediction while its parent continues toward full-depth verification. Shared-core batches advance lanes at different positions and recurrent depths. Optional $q_2$ reads stage $X$, at depth $XK$ rather than necessarily $2K$. Its greedy fallback gate opens when $g_X(c)\ne g_1(c)$. For sampling, given a first proposal $y\sim p_1(\cdot\mid c)$, the gate opens when $p_X(y\mid c)<p_1(y\mid c)$. The fallback is drawn from the residual proposal
\begin{equation}
 R(x\mid c)=
 \frac{[p_X(x\mid c)-p_1(x\mid c)]_+}
      {\sum_v[p_X(v\mid c)-p_1(v\mid c)]_+}.
 \label{eq:baseline-loopspec-residual}
\end{equation}
The sampling gate ensures a positive denominator. After first-proposal rejection, let $F=p_{\mathrm{rej}}(\cdot\mid c)$ from Eq.~\ref{eq:speculative-verification} with $p=p_P$ and $q=p_1$. A fallback draw $y'\sim R$ is accepted with probability $\min\{1,F(y')/R(y')\}$. Final selection preserves the chosen child's progress and prunes incompatible descendants. Each proposal starts one token input, rather than a multi-input block.

\paragraph{WaveFront Decoding (WFD).}
WFD~\citep{ha2026wavefront} also designs a self-speculative pipeline specifically for looped language models. It organizes positions at different recurrent depths into a diagonal wavefront. Shallow readouts introduce new positions while shared batched core calls advance earlier ones toward final verification. Under the $K\mid S$ schedules used here, ideal steady-state operation with uninterrupted acceptance gives
\begin{equation}
 |\mathcal{W}|=P,\qquad
 L_{\mathrm{commit}}=\frac{N_{\mathrm{decode}}}{C}=1.
 \label{eq:baseline-wavefront-steady}
\end{equation}
Here $\mathcal{W}$ is the active wavefront. $N_{\mathrm{decode}}$ and $C$ count final tokens and $K$-iteration cycles within a steady-state interval, excluding warmup and drain. This is a scheduling relation, not a wall-clock speedup. A rejection commits the correction, flushes later speculative positions, and regrows the wavefront. Our reference ports the public Transformers scheduler to SGLang. Table~\ref{tab:main-performance} reports the comparison. The paper separately studies optional cross-recurrence KV sharing.

\subsection{Candidate Budgets and Tree Admission}
\label{app:tree-budget-admission}

\paragraph{Candidate sources and allocation.}
For a tree node with context $c$, the builder queries history continuations and the available $q_1$ candidate view. Full-input computed rows, when enabled, use a separate provisional index and take precedence over bounded-context caches. Matching branch-local rows precede committed rows at the same key. Shorter suffixes are proposal fallbacks, subject to the configured length and support checks. A cached top-$k$ list need not fit in the tree: node and depth limits may exclude some of its candidates.

The spine layout protects a leading continuation, following $q_1$ top-1 where available and otherwise an admissible historical edge. Remaining slots are reserved for alternatives under ancestor closure: a retained node must retain its complete root path. The implementation supports breadth-oriented and root/early-boundary repair allocations, with the policy chosen per model. CPU construction may consider a larger union, but prunes it to the verified-node budget before graph submission. An automaton candidate may therefore be absent from the submitted tree, and only selected, computed rows supply reusable $q_1$ states.

Writing the retrieved root-anchored paths as $\mathcal{C}(r)$, prefix merging and pruning give
\begin{equation}
 \begin{gathered}
 \mathcal{T}=\operatorname{Prune}_{N,D}
                \bigl(\operatorname{Trie}(\mathcal{C}(r))\bigr),\\
 |\mathcal{T}|\le N,\qquad \max_{v\in\mathcal{T}}d(v)<D.
 \end{gathered}
 \label{eq:draft-tree-budget}
\end{equation}
The root has depth zero, so $D$ counts inputs including the root. Duplicate prefixes receive one row.

\paragraph{Candidate publication.}
The memory $\mathcal{H}$ stores history windows with occurrence counts, and $\mathcal{M}_g$ stores $q_1$ candidates for committed decisions. For a $q_1$ readout under context $c$, write $L(c)=\operatorname{Top}_k(z_1(c))$. Equation~\ref{eq:q1-publication} publishes this list to the current branch when the $q_1$ result is consumed. GPU completion alone does not publish it. Upon commitment of the final prediction $y$ under the same context,
\begin{equation}
 \begin{aligned}
 \mathcal{M}_g[\kappa_n(c)]&\leftarrow L(c),\\
 \mathcal{H}&\leftarrow\mathcal{H}\uplus\Delta_n(c,y).
 \end{aligned}
 \label{eq:committed-memory-update}
\end{equation}
Here $\Delta_n(c,y)$ contains the newly completed history windows ending in $y$, and $\uplus$ preserves occurrence counts. Promotion keeps the original $q_1$ row rather than replacing it with final logits. Suffix-key matches supply candidate tokens, not evidence that the current context's logits or hidden/KV have already been computed.

\paragraph{Yield-based admission.}
For each root token $r$, the gate tracks selected $q_1$ path length $m$, including the anchored root. After its first observation, it updates
\begin{equation}
 \bar m_r\leftarrow\tfrac12\bar m_r+\tfrac12 m.
 \label{eq:tree-yield-gate}
\end{equation}
By default, the first two observations are allowed. Thereafter, roots with $\bar m_r<\tau$ skip tree work, except for a probe on every eighth otherwise-skipped eligible check. These probes let newly accumulated evidence recover from an earlier low-yield period. Missing history may be supplemented by a sufficiently long logits-derived chain when that option is enabled. Prefix readiness, capacity, warmup, and configured ancestry checks still apply. The gate admits speculative work, while final verification remains unchanged. Optional measured-cost policies are model-specific.

\subsection{Complete Schedule and Cross-Model Stages}
\label{app:complete-schedule}

Algorithms~\ref{alg:block-wavefront}--\ref{alg:wavefront-continuation} give the greedy schedule, $q_1$ feedback, and final continuation. Full-model prefill populates the prompt's role-specific KV and emits the first completion token. The committed prefix $c$ includes that token, and $\mathcal{W}$ initially contains its input lane at depth zero. The memory $\mathcal{A}=(\mathcal{H},\mathcal{M}_g,\{\mathcal{M}_b\})$ is initialized with prompt/completion history and empty logit views. Each input $u$ records its complete context $c(u)$, completed depth, native state/KV, saved $q_1$ prediction $d_1(u)$, and original top-$k$ list $L_u$.

\paragraph{Scheduling and state ownership.}
Algorithm~\ref{alg:block-wavefront} advances dependency-ready rows in $K$-iteration stages. Replacing an ordinary $q_1$ job by a tree removes duplicate root work. \textsc{LaunchAndCollect} updates completed states and readouts in $\mathcal{W}$ and returns newly consumable results. It may share supported core batches without launching all jobs in one graph. Producer events and live ancestry are checked before consumption. Completed rows that cannot yet be installed retain ownership, and \textsc{InstallPending} admits them when capacity and dependencies permit. Same-stage ancestors must be present in the actual causal batch or in ready prefix KV.

\begin{algorithm}[t]
\caption{Greedy block-wavefront decoding}
\label{alg:block-wavefront}
\small
\begin{algorithmic}[1]
\Require Prompt $x$, schedule $(S,K,X)$, budgets $N,D$
\Require Top-$k$ width $k$, context-key length $n$
\State $(c,\mathrm{KV})\gets\Call{FullPrefill}{x,S}$
\State $\mathcal{A}\gets\Call{InitMemory}{c}$
\State $\mathcal{W}\gets\{\Call{NewLane}{c,0}\}$
\While{not $\Call{Stop}{c}$}
  \State \Call{InstallPending}{$\mathcal{W}$}
  \State $\mathcal{J}\gets\Call{DependencyReadyWork}{\mathcal{W}}$
  \State $u\gets\Call{SelectTreeRoot}{\mathcal{W}}$
  \If{$u\ne\emptyset$ and $\Call{Admit}{u,\mathcal{A}}$}
    \State $\mathcal{C}\gets\Call{RetrievePaths}{u,\mathcal{A}}$
    \State $\mathcal{T}\gets\operatorname{Prune}_{N,D}(\operatorname{Trie}(\mathcal{C}))$
    \State $\mathcal{J}\gets\Call{ReplaceQ1Work}{\mathcal{J},u,\mathcal{T}}$
  \EndIf
  \State $\mathcal{E}\gets\Call{LaunchAndCollect}{\mathcal{J},K,\mathcal{W}}$
  \ForAll{newly consumed valid $q_1$ results $e\in\mathcal{E}$}
    \State \Call{ConsumeQ1}{$e,\mathcal{W},\mathcal{A}$}
  \EndFor
  \State \Call{OfferQ2}{$\mathcal{E},\mathcal{W},\mathcal{A}$}
  \State $c\gets\Call{CommitReady}{\mathcal{W},\mathcal{A},c}$
\EndWhile
\State \Return generated suffix of $c$
\end{algorithmic}
\end{algorithm}

\paragraph{Tree selection and early feedback.}
Algorithm~\ref{alg:q1-feedback} follows only matching $q_1$ edges and publishes each selected input's saved top-$k$ row under its own context key. Ordinary $q_1$ and completed bonus inputs are singleton trees. Readout selection uses the completed rows, not a new forward per node. \textsc{BindQ1} attaches each selected row to its matching input at depth $K$, or queues installation without recomputing it. A compatible input already farther ahead retains its progress. \textsc{InheritView} merges ancestor-local evidence with the child's own compatible rows, without exposing rejected siblings. It is a no-op for a root without a parent. \textsc{CanExtend} enforces the remaining token and context limits. The last selected row updates the key for $c_m$. Its bonus input at $c_m\mathbin{\Vert}\beta$ contributes no new logits until its own $q_1$ computation completes.

\begin{algorithm}[t]
\caption{$q_1$ tree selection and feedback}
\label{alg:q1-feedback}
\small
\begin{algorithmic}[1]
\Require Consumed $q_1$ result $e$ with live owner, $\mathcal{W},\mathcal{A}$
\Procedure{ConsumeQ1}{$e,\mathcal{W},\mathcal{A}$}
  \State Read $\mathcal{T}$ and completed rows $(H_v,\mathsf{C}_v,z_v)$ from $e$
  \State For all real nodes $v$, obtain $\hat y_v=\arg\max_x z_v[x]$
  \State For all real nodes $v$, obtain $L_v=\operatorname{Top}_k(z_v)$
  \State $v\gets\Call{Root}{\mathcal{T}}$
  \State $w\gets\Call{ParentOfRoot}{e}$
  \While{$v\ne\emptyset$}
    \State $u\gets\Call{BindQ1}{\mathcal{W},w,v,H_v,\mathsf{C}_v}$
    \State \Call{InheritView}{$\mathcal{A},w,u$}
    \State Save $L_u\gets L_v$ and $d_1(u)\gets\hat y_v$
    \State $\mathcal{M}_u[\kappa_n(c(u))]\gets L_u$
    \State $w\gets u$
    \State $v\gets\Call{MatchingChild}{\mathcal{T},v,\hat y_v}$
  \EndWhile
  \State $\beta\gets d_1(u)$
  \If{$\Call{CanExtend}{u}$}
    \State $u^{+}\gets\Call{ReuseOrNew}{\mathcal{W},u,\beta}$
    \State \Call{InheritView}{$\mathcal{A},u,u^{+}$}
  \EndIf
\EndProcedure
\end{algorithmic}
\end{algorithm}

\paragraph{Alternative selection and continuous commitment.}
Algorithm~\ref{alg:wavefront-continuation} first offers admitted $q_2$ alternatives without pruning $q_1$. \textsc{AdmitAlt} applies the configured capacity and work-admission checks. \textsc{ReuseOrNew} attaches a live or cached successor under the full context $c(u)\mathbin{\Vert}y$, using the matching state contract in Eq.~\ref{eq:continuation-depth}, and restores compatible descendants when available. Otherwise it creates a depth-zero input. It inherits no hidden/KV from a different continuation. During commitment, \textsc{FrontierFinalReady} requires $c(u)=c$ and a completed final readout. After emitting $y$, the algorithm promotes the original $L_u$ and inserts the new history windows. \textsc{RetainContinuation} preserves the selected successor and its compatible descendants. Pruning removes discarded branches' local views, not committed memory. The loop immediately consumes another ready FINAL on the selected path, without launching new target work. EOS and length limits are checked after each emitted token.

\begin{algorithm}[t]
\caption{Alternative successors and ordered commitment}
\label{alg:wavefront-continuation}
\small
\begin{algorithmic}[1]
\Procedure{OfferQ2}{$\mathcal{E},\mathcal{W},\mathcal{A}$}
  \ForAll{inputs $u$ with a new valid $q_2$ readout in $\mathcal{E}$}
    \State $y\gets g_X(c(u))$
    \If{$y\ne d_1(u)$ and $\Call{AdmitAlt}{u,y}$}
      \State $v\gets\Call{ReuseOrNew}{\mathcal{W},u,y}$
      \State \Call{InheritView}{$\mathcal{A},u,v$}
      \State Keep both $q_1$ and $q_2$ successors of $u$
    \EndIf
  \EndFor
\EndProcedure
\Statex
\Procedure{CommitReady}{$\mathcal{W},\mathcal{A},c$}
  \While{$\Call{FrontierFinalReady}{\mathcal{W},c}$}
    \State $u\gets\Call{FrontierLane}{\mathcal{W},c}$
    \State $y\gets g_P(c(u))$
    \State $c\gets c\mathbin{\Vert}y$
    \If{the original $q_1$ list $L_u$ is available}
      \State $\mathcal{M}_g[\kappa_n(c(u))]\gets L_u$
    \EndIf
    \State $\mathcal{H}\gets\mathcal{H}\uplus\Delta_n(c(u),y)$
    \If{$\Call{Stop}{c}$}
      \State \Return $c$
    \EndIf
    \State $v\gets\Call{ReuseOrNew}{\mathcal{W},u,y}$
    \State \Call{InheritView}{$\mathcal{A},u,v$}
    \State \Call{RetainContinuation}{$\mathcal{W},\mathcal{A},v$}
  \EndWhile
  \State \Return $c$
\EndProcedure
\end{algorithmic}
\end{algorithm}

\paragraph{Proposal endpoints and recurrent stages.}
The labels $q_1$ and $q_2$ denote proposal endpoints, whereas $F_j$ and $p_j$ index stages. The optional $q_2$ endpoint is at $XK$ iterations, and its candidate does not prune $q_1$ before final selection. The table below lists the controlled-comparison schedules. Prelude and coda work is additional to the recurrent-stage count.
\begin{center}
\small
\setlength{\tabcolsep}{3pt}
\begin{tabular}{@{}lrrrr@{}}
\toprule
Checkpoints & $S$ & $K$ & $P$ & $q_2$ depth \\
\midrule
Ouro / Thinking & 4 & 1 & 4 & 2 \\
Huginn / Llama / Tiny & 32 & 8 & 4 & 16 \\
Recurrent-OLMo-2 & 32 & 8 & 4 & 16 \\
\bottomrule
\end{tabular}
\end{center}
The stage count is $S/K$. All eight checkpoints preserve their full target recurrence. Model-specific prelude outputs, injected states, and coda KV remain part of their own execution and cache contracts.

\paragraph{Continuation example.}
Suppose $q_1$ selected $(r,a,b)$, and its ordinary final rows predict $(a,b,c)$. Once $r$ extends the committed prefix, the scheduler can emit $a,b,c$ consecutively if all three rows are ready. A matching input $c$ already at depth $2K$ continues from $2K$, not zero. If the row for $a$ instead predicts $d\ne b$, only $a,d$ are emitted, and the $b$ subtree is discarded. A precomputed $d$ successor can retain progress only under the exact prefix ending in $(r,a,d)$ and its validated cache provenance. $q_2$ agreement alone is insufficient. Figure~\ref{fig:branch-pipeline} expands this example across consecutive blocks and a rejected alternative branch.

\subsection{Cache Identity and Sampling}

\paragraph{Proposal evidence versus reusable computation.}
Automaton history, committed $q_1$ rows, and branch-local rows are request-local proposal sources. Suffix matches supply proposal candidates. Hidden/KV reuse requires the actual input token, absolute position, complete retained context, recurrent stage, cache role, and physical ownership to match. For configurations with a read-only committed FINAL-prefix overlay in the $q_1$ readout, the producer's read bound is also part of the identity. This optional overlay changes only the shallow coda readout. Recurrent core caches remain stage-specific. Its read bound is frozen at launch and never includes the current input or an uncomputed bonus. Only ordered, completed FINAL inputs are published. Pending graph outputs remain owned until consumption or cancellation, with dependencies satisfied before a consumer reads or installs them. 

\paragraph{Chain visibility and continuation.}
Index a dependency-ready segment from one, as in the main text. Restricting tree attention to this chain gives the within-segment mask
\begin{equation}
 M_{\triangle}(i,h)=
 \begin{cases}
 0,&1\le h\le i,\\
 -\infty,&h>i.
 \end{cases}
 \label{eq:chain-causal-mask}
\end{equation}
All matching prefix keys remain visible. At every executed iteration and attention role, reads wait for their producers or use compatible ancestor rows in the same causal batch.

After Eq.~\ref{eq:final-prefix-commit}, let $c_{\mathrm{new}}=c_1\mathbin{\Vert}\Delta_{\mathrm{out}}$. Its last token is the next input. For a candidate successor $v$ at depth $s(v)$, the retained depth is
\begin{equation}
 s_{\mathrm{next}}=
 \begin{cases}
 s(v),&\operatorname{Match}\bigl(v,c_{\mathrm{new}}\bigr),\\
 0,&\text{otherwise}.
 \end{cases}
 \label{eq:continuation-depth}
\end{equation}
$\operatorname{Match}$ requires the complete context, absolute position, native state provenance, and role-specific KV ownership and read bounds to agree. It is false if no eligible successor exists. Compatible descendants retain their progress, and incompatible descendants are pruned.

\paragraph{Stochastic tree selection.}
For temperature-based decoding, each real tree input receives one draw from its own filtered $q_1$ distribution. The selected path follows these draws through unique token-prefix edges. An absent child ends the path, with the drawn token becoming its next proposal. Duplicate copies of the same prefix with independent draws are disallowed, as choosing a lucky continuation would bias selection. Each retained input owns the distribution that generated its child proposal. Neither automaton frequencies nor top-$k$ ranks substitute for that distribution.

\paragraph{Rejection-based acceptance and fallback.}
Let $q$ be the normalized proposal distribution actually used to draw $y$, and $p$ the normalized final distribution obtained from final logits with the configured temperature and sampling filters. Both refer to the same causal context $c$. The first proposal uses Eq.~\ref{eq:speculative-verification}. On rejection, the conditional target is $F=p_{\mathrm{rej}}(\cdot\mid c)$. An optional $q_2$ alternative is drawn from its saved residual proposal $R$, not assumed to come directly from the $q_2$ readout. It is accepted against $F$ using $\min\{1,F(y)/R(y)\}$. If it also rejects, fallback samples $\operatorname{normalize}((F-R)_+)$. Without a $q_2$ alternative, fallback samples $F$. Residuals are not top-$p$ filtered again. Verification is consumed in prefix order, and a rejected prefix invalidates its speculative descendants. Residual normalization is required only on positive-probability rejection events.

\subsection{\texorpdfstring{Work Accounting and Correctness}{Work Accounting and Correctness}}
\label{app:method-work}

\paragraph{Recurrent work and throughput.}
Let $W\ge N'$ be the physical graph width for a tree of $N'\le N$ real nodes, including any padding, and $m$ its selected path length. For a fresh tree whose entire selected path survives final verification, assuming exactly $m$ rows in the remaining-stage batches, an idealized count of recurrent token-iterations is
\begin{equation}
 \begin{aligned}
 R_{\mathrm{core}} &= K\,W+(S-K)m\\
                 &= S\,m+K(W-m).
 \end{aligned}
 \label{eq:method-core-work}
\end{equation}
Here $Sm$ is the work for those $m$ inputs without the tree's unused rows. This count excludes prelude/coda work, repair branches, reused anchors, and rejected ancestors. Batched execution amortizes this recurrent work across multiple inputs. Latency also depends on graph width, context length, and synchronization. In particular, committed decode throughput decomposes exactly as
\begin{equation}
 \mathrm{TPS}=\frac{N_{\mathrm{decode}}}{C}
                  \cdot\frac{C}{t_{\mathrm{decode}}}.
 \label{eq:method-throughput-decomposition}
\end{equation}
Ancestor survival and continuation determine tokens committed per cycle, while uncovered tree and repair work can increase time per cycle. Local acceptance alone therefore cannot predict throughput. The experimental counters report both local path length and final non-root inputs realized per tree call.

\paragraph{Equivalence argument.}
Assume exact arithmetic, matching native initialization and frozen readout metadata, and that each reused state and role-specific KV equals its ordinary per-context counterpart. At each attention role, Eqs.~\ref{eq:tree-causal-mask} and~\ref{eq:chain-causal-mask} expose exactly the input's causal prefix, including itself. The same per-row operations therefore produce the same hidden update and KV writes. Induction over recurrent iterations and token positions gives the ordinary final logits for each surviving input. Equation~\ref{eq:final-prefix-commit} stops at the first incompatible successor, so no later decision under that rejected context is emitted. Equation~\ref{eq:continuation-depth} permits reuse only under a matching execution contract. For sampling, the stored proposal distribution and residual corrections above are additional requirements.

Appendix~\ref{app:sanity-checks} reports numerical checks and greedy task-quality evaluations.

\FloatBarrier
\section{Use of AI Assistants}
\label{app:ai-assistance}

AI assistants were used to help debug implementation code and edit the manuscript's language and presentation. The research ideas, method design, experimental protocols, and interpretation of results were determined by the authors. The authors are responsible for the final implementation and manuscript.


\begin{thebibliography}{25}
\providecommand{\natexlab}[1]{#1}

\bibitem[{Cai et~al.(2024)Cai, Li, Geng, Peng, Lee, Chen, and
  Dao}]{cai2024medusa}
Tianle Cai, Yuhong Li, Zhengyang Geng, Hongwu Peng, Jason~D. Lee, Deming Chen,
  and Tri Dao. 2024.
\newblock \href {https://proceedings.mlr.press/v235/cai24b.html} {Medusa:
  Simple {LLM} inference acceleration framework with multiple decoding heads}.
\newblock In \emph{Proceedings of the 41st International Conference on Machine
  Learning}, volume 235 of \emph{Proceedings of Machine Learning Research},
  pages 5209--5235. PMLR.

\bibitem[{Chen et~al.(2023)Chen, Borgeaud, Irving, Lespiau, Sifre, and
  Jumper}]{chen2023speculative}
Charlie Chen, Sebastian Borgeaud, Geoffrey Irving, Jean-Baptiste Lespiau,
  Laurent Sifre, and John Jumper. 2023.
\newblock \href {https://arxiv.org/abs/2302.01318} {Accelerating large language
  model decoding with speculative sampling}.
\newblock \emph{arXiv preprint arXiv:2302.01318}.

\bibitem[{Cho et~al.(2026)Cho, Cui, Kim, Han, and Han}]{cho2026loopspec}
SangLyul Cho, Langqing Cui, Sehoon Kim, Dongsu Han, and Insu Han. 2026.
\newblock \href {https://arxiv.org/abs/2609.17184} {{LoopSpec}: Pipelined
  self-speculative decoding for looped transformers}.
\newblock \emph{arXiv preprint arXiv:2609.17184}.

\bibitem[{Cobbe et~al.(2021)Cobbe, Kosaraju, Bavarian, Chen, Jun, Kaiser,
  Plappert, Tworek, Hilton, Nakano, Hesse, and Schulman}]{cobbe2021gsm8k}
Karl Cobbe, Vineet Kosaraju, Mohammad Bavarian, Mark Chen, Heewoo Jun, Lukasz
  Kaiser, Matthias Plappert, Jerry Tworek, Jacob Hilton, Reiichiro Nakano,
  Christopher Hesse, and John Schulman. 2021.
\newblock \href {https://arxiv.org/abs/2110.14168} {Training verifiers to solve
  math word problems}.
\newblock \emph{arXiv preprint arXiv:2110.14168}.

\bibitem[{Elhoushi et~al.(2024)Elhoushi, Shrivastava, Liskovich, Hosmer, Wasti,
  Lai, Mahmoud, Acun, Agarwal, Roman, Aly, Chen, and
  Wu}]{elhoushi2024layerskip}
Mostafa Elhoushi, Akshat Shrivastava, Diana Liskovich, Basil Hosmer, Bram
  Wasti, Liangzhen Lai, Anas Mahmoud, Bilge Acun, Saurabh Agarwal, Ahmed Roman,
  Ahmed Aly, Beidi Chen, and Carole-Jean Wu. 2024.
\newblock \href {https://doi.org/10.18653/v1/2024.acl-long.681} {{LayerSkip}:
  Enabling early exit inference and self-speculative decoding}.
\newblock In \emph{Proceedings of the 62nd Annual Meeting of the Association
  for Computational Linguistics (Volume 1: Long Papers)}, pages 12622--12642.
  Association for Computational Linguistics.

\bibitem[{Geiping et~al.(2025)Geiping, McLeish, Jain, Kirchenbauer, Singh,
  Bartoldson, Kailkhura, Bhatele, and Goldstein}]{geiping2025latent}
Jonas Geiping, Sean McLeish, Neel Jain, John Kirchenbauer, Siddharth Singh,
  Brian Bartoldson, Bhavya Kailkhura, Abhinav Bhatele, and Tom Goldstein. 2025.
\newblock \href {https://doi.org/10.52202/085713-1380} {Scaling up test-time
  compute with latent reasoning: A recurrent depth approach}.
\newblock In \emph{Advances in Neural Information Processing Systems},
  volume~38.

\bibitem[{Ha and Kim(2026)}]{ha2026wavefront}
Hyeongju Ha and Jae-Joon Kim. 2026.
\newblock \href {https://arxiv.org/abs/2609.23033} {{WaveFront} decoding:
  Parallelized self-speculative decoding for looped language models}.
\newblock \emph{arXiv preprint arXiv:2609.23033}.

\bibitem[{He et~al.(2024)He, Zhong, Cai, Lee, and He}]{he2024rest}
Zhenyu He, Zexuan Zhong, Tianle Cai, Jason Lee, and Di~He. 2024.
\newblock \href {https://doi.org/10.18653/v1/2024.naacl-long.88} {{REST}:
  Retrieval-based speculative decoding}.
\newblock In \emph{Proceedings of the 2024 Conference of the North American
  Chapter of the Association for Computational Linguistics: Human Language
  Technologies (Volume 1: Long Papers)}, pages 1582--1595. Association for
  Computational Linguistics.

\bibitem[{Hendrycks et~al.(2021)Hendrycks, Burns, Kadavath, Arora, Basart,
  Tang, Song, and Steinhardt}]{hendrycks2021math}
Dan Hendrycks, Collin Burns, Saurav Kadavath, Akul Arora, Steven Basart, Eric
  Tang, Dawn Song, and Jacob Steinhardt. 2021.
\newblock \href {https://arxiv.org/abs/2103.03874} {Measuring mathematical
  problem solving with the {MATH} dataset}.
\newblock \emph{arXiv preprint arXiv:2103.03874}.

\bibitem[{Hooper et~al.(2023)Hooper, Kim, Mohammadzadeh, Genc, Keutzer,
  Gholami, and Shao}]{hooper2023speed}
Coleman Hooper, Sehoon Kim, Hiva Mohammadzadeh, Hasan Genc, Kurt Keutzer, Amir
  Gholami, and Yakun~Sophia Shao. 2023.
\newblock \href {https://arxiv.org/abs/2310.12072} {{SPEED}: Speculative
  pipelined execution for efficient decoding}.
\newblock \emph{arXiv preprint arXiv:2310.12072}.

\bibitem[{Leviathan et~al.(2023)Leviathan, Kalman, and
  Matias}]{leviathan2023speculative}
Yaniv Leviathan, Matan Kalman, and Yossi Matias. 2023.
\newblock \href {https://proceedings.mlr.press/v202/leviathan23a.html} {Fast
  inference from transformers via speculative decoding}.
\newblock In \emph{Proceedings of the 40th International Conference on Machine
  Learning}, volume 202 of \emph{Proceedings of Machine Learning Research},
  pages 19274--19286. PMLR.

\bibitem[{Li et~al.(2026)Li, Su, Yu, Jiang, Ren, Li, Lu, Tang, and
  Yuan}]{li2026daswave}
Guanghao Li, Zihan Su, Hao Yu, Jinyang Jiang, Tao Ren, Zehao Li, Feng Lu, Ming
  Tang, and Chun Yuan. 2026.
\newblock \href {https://arxiv.org/abs/2609.34538} {Shallow queries, mature
  values: Depth-asynchronous self-speculation for looped transformers}.
\newblock \emph{arXiv preprint arXiv:2609.34538}.

\bibitem[{Li et~al.(2024)Li, Wei, Zhang, and Zhang}]{li2024eagle}
Yuhui Li, Fangyun Wei, Chao Zhang, and Hongyang Zhang. 2024.
\newblock \href {https://proceedings.mlr.press/v235/li24bt.html} {{EAGLE}:
  Speculative sampling requires rethinking feature uncertainty}.
\newblock In \emph{Proceedings of the 41st International Conference on Machine
  Learning}, volume 235 of \emph{Proceedings of Machine Learning Research},
  pages 28935--28948. PMLR.

\bibitem[{Lightman et~al.(2023)Lightman, Kosaraju, Burda, Edwards, Baker, Lee,
  Leike, Schulman, Sutskever, and Cobbe}]{lightman2023verify}
Hunter Lightman, Vineet Kosaraju, Yura Burda, Harri Edwards, Bowen Baker, Teddy
  Lee, Jan Leike, John Schulman, Ilya Sutskever, and Karl Cobbe. 2023.
\newblock \href {https://arxiv.org/abs/2305.20050} {Let's verify step by step}.
\newblock \emph{arXiv preprint arXiv:2305.20050}.

\bibitem[{Liu et~al.(2023)Liu, Xia, Wang, and Zhang}]{liu2023evalplus}
Jiawei Liu, Chunqiu~Steven Xia, Yuyao Wang, and Lingming Zhang. 2023.
\newblock \href {https://arxiv.org/abs/2305.01210} {Is your code generated by
  {ChatGPT} really correct? rigorous evaluation of large language models for
  code generation}.
\newblock \emph{arXiv preprint arXiv:2305.01210}.

\bibitem[{Luo et~al.(2025)Luo, Wang, Zhu, Zhang, Zhang, Yang, and
  Xu}]{luo2025recycling}
Xianzhen Luo, Yixuan Wang, Qingfu Zhu, Zhiming Zhang, Xuanyu Zhang, Qing Yang,
  and Dongliang Xu. 2025.
\newblock \href {https://doi.org/10.18653/v1/2025.acl-long.338} {Turning trash
  into treasure: Accelerating inference of large language models with token
  recycling}.
\newblock In \emph{Proceedings of the 63rd Annual Meeting of the Association
  for Computational Linguistics (Volume 1: Long Papers)}, pages 6816--6831.
  Association for Computational Linguistics.

\bibitem[{Ma et~al.(2026)Ma, Zhang, Li, Zhang, Qi, Li, and Yu}]{ma2026echo}
Ziyang Ma, Zihong Zhang, Zuchao Li, Lefei Zhang, Baoyuan Qi, Siqi Li, and Simin
  Yu. 2026.
\newblock \href {https://arxiv.org/abs/2609.17241} {{ECHO}: Early-layer
  collaborative hierarchical orchestration with bonus logits in speculative
  decoding}.
\newblock \emph{arXiv preprint arXiv:2609.17241}.

\bibitem[{Marzollo et~al.(2026)Marzollo, Zhuang, Roemer, Zwingenberger, Muller,
  and Cavigelli}]{marzollo2026sssd}
Michele Marzollo, Jiawei Zhuang, Niklas Roemer, Niklas Zwingenberger, Lorenz~K
  Muller, and Lukas Cavigelli. 2026.
\newblock \href {https://doi.org/10.18653/v1/2026.acl-long.1530} {{SSSD}:
  Simply-scalable speculative decoding}.
\newblock In \emph{Proceedings of the 64th Annual Meeting of the {A}ssociation
  for {C}omputational {L}inguistics (Volume 1: Long Papers)}, pages
  33123--33139, San Diego, California, United States. Association for
  Computational Linguistics.

\bibitem[{McDanel et~al.(2025)McDanel, Zhang, Hu, and
  Liu}]{mcdanel2025pipespec}
Bradley McDanel, Sai~Qian Zhang, Yunhai Hu, and Zining Liu. 2025.
\newblock \href {https://doi.org/10.18653/v1/2025.findings-acl.669}
  {{PipeSpec}: Breaking stage dependencies in hierarchical {LLM} decoding}.
\newblock In \emph{Findings of the Association for Computational Linguistics:
  ACL 2025}, pages 12909--12920. Association for Computational Linguistics.

\bibitem[{Suzgun et~al.(2022)Suzgun, Scales, Sch{\"a}rli, Gehrmann, Tay, Chung,
  Chowdhery, Le, Chi, Zhou, and Wei}]{suzgun2022bbh}
Mirac Suzgun, Nathan Scales, Nathanael Sch{\"a}rli, Sebastian Gehrmann, Yi~Tay,
  Hyung~Won Chung, Aakanksha Chowdhery, Quoc~V. Le, Ed~H. Chi, Denny Zhou, and
  Jason Wei. 2022.
\newblock \href {https://arxiv.org/abs/2210.09261} {Challenging {BIG-Bench}
  tasks and whether chain-of-thought can solve them}.
\newblock \emph{arXiv preprint arXiv:2210.09261}.

\bibitem[{Yu et~al.(2026)Yu, Wang, Yuan, Ren, and Pei}]{yu2026spd}
Yijiong Yu, Huazheng Wang, Shuai Yuan, Ruilong Ren, and Ji~Pei. 2026.
\newblock \href {https://arxiv.org/abs/2605.30852} {Speculative pipeline
  decoding: Higher-accuracy drafting with hidden latency via pipeline
  parallelism}.
\newblock \emph{arXiv preprint arXiv:2605.30852}.

\bibitem[{Zhang et~al.(2024)Zhang, Wang, Li, Shou, Chen, Chen, and
  Mehrotra}]{zhang2024draftverify}
Jun Zhang, Jue Wang, Huan Li, Lidan Shou, Ke~Chen, Gang Chen, and Sharad
  Mehrotra. 2024.
\newblock \href {https://doi.org/10.18653/v1/2024.acl-long.607} {Draft {\&}
  verify: Lossless large language model acceleration via self-speculative
  decoding}.
\newblock In \emph{Proceedings of the 62nd Annual Meeting of the Association
  for Computational Linguistics (Volume 1: Long Papers)}, pages 11263--11282.
  Association for Computational Linguistics.

\bibitem[{Zhang et~al.(2026)Zhang, Li, Zhang, Wang, and Zhao}]{zhang2026racer}
Zihong Zhang, Zuchao Li, Lefei Zhang, Ping Wang, and Hai Zhao. 2026.
\newblock \href {https://doi.org/10.18653/v1/2026.findings-acl.998} {{RACER}:
  Retrieval-augmented contextual rapid speculative decoding}.
\newblock In \emph{Findings of the Association for Computational Linguistics:
  ACL 2026}, pages 19962--19988. Association for Computational Linguistics.

\bibitem[{Zheng et~al.(2023)Zheng, Yin, Xie, Sun, Huang, Yu, Cao, Kozyrakis,
  Stoica, Gonzalez, Barrett, and Sheng}]{zheng2023sglang}
Lianmin Zheng, Liangsheng Yin, Zhiqiang Xie, Chuyue Sun, Jeff Huang, Cody~Hao
  Yu, Shiyi Cao, Christos Kozyrakis, Ion Stoica, Joseph~E. Gonzalez, Clark
  Barrett, and Ying Sheng. 2023.
\newblock \href {https://arxiv.org/abs/2312.07104} {{SGLang}: Efficient
  execution of structured language model programs}.
\newblock \emph{arXiv preprint arXiv:2312.07104}.

\bibitem[{Zhu et~al.(2025)Zhu, Wang, Hua, Zhang, Li, Que, Wei, Wen, Yin, Xing,
  Li, Shi, Ma, Li, Kergan, Smith, Qu, Hui, Wu, Min, Huang, Zhou, Ye, Liu, Yang,
  Shi, Lin, Zhao, Cai, Zhang, Huang, Bengio, and Eshraghian}]{zhu2025ouro}
Rui-Jie Zhu, Zixuan Wang, Kai Hua, Tianyu Zhang, Ziniu Li, Haoran Que, Boyi
  Wei, Zixin Wen, Fan Yin, He~Xing, Lu~Li, Jiajun Shi, Kaijing Ma, Shanda Li,
  Taylor Kergan, Andrew Smith, Xingwei Qu, Mude Hui, Bohong Wu, and 14 others.
  2025.
\newblock \href {https://arxiv.org/abs/2510.25741} {Scaling latent reasoning
  via looped language models}.
\newblock \emph{arXiv preprint arXiv:2510.25741}.

\end{thebibliography}
\end{document}